\documentclass[fleqn,10pt]{wlscirep}
\usepackage[utf8]{inputenc}
\usepackage[T1]{fontenc}
\usepackage{lineno}
\usepackage{float}
\usepackage{hyperref}
\restylefloat{table}

\title{{\fontsize{19}{22}\selectfont A word association network methodology for evaluating implicit biases in LLMs compared to humans}}

\author[1,*]{Katherine Abramski}
\author[2,+]{Giulio Rossetti}
\author[3,+]{Massimo Stella}
\affil[1]{University of Pisa, Department of Computer Science, Pisa, Italy}
\affil[2]{National Research Council of Italy, Institute of Information Science and Technologies, Pisa, Italy}
\affil[3]{University of Trento, Department of Psychology and Cognitive Science, Trento, Italy}
\affil[*]{Corresponding author: Katherine Abramski (katherine.abramski@phd.unipi.it).}
\affil[+]{These authors contributed equally}

\begin{abstract}
As Large language models (LLMs) become increasingly integrated into our lives, their inherent social biases remain a pressing concern. Detecting and evaluating these biases can be challenging because they are often implicit rather than explicit in nature, so developing evaluation methods that assess the implicit knowledge representations of LLMs is essential. We present a novel word association network methodology for evaluating implicit biases in LLMs based on simulating semantic priming within LLM-generated word association networks. Our prompt-based approach taps into the implicit relational structures encoded in LLMs, providing both quantitative and qualitative assessments of bias. Unlike most prompt-based evaluation methods, our method enables direct comparisons between various LLMs and humans, providing a valuable point of reference and offering new insights into the alignment of LLMs with human cognition. To demonstrate the utility of our methodology, we apply it to both humans and several widely used LLMs to investigate social biases related to gender, religion, ethnicity, sexual orientation, and political party. Our results reveal both convergences and divergences between LLM and human biases, providing new perspectives on the potential risks of using LLMs. Our methodology contributes to a systematic, scalable, and generalizable framework for evaluating and comparing biases across multiple LLMs and humans, advancing the goal of transparent and socially responsible language technologies.

\end{abstract}
\begin{document}

\flushbottom
\maketitle

\thispagestyle{empty}

\section{Introduction}
Large Language Models (LLMs) are rapidly transforming our lives by revolutionizing the way we produce and consume digital content. Their accessibility and versatility have led to widespread use across diverse domains and NLP tasks—from writing emails and debugging code to making high-stakes decisions and participating in online discussions that influence opinion dynamics \cite{cau2025selective}. Although LLMs hold great potential to enhance human productivity and understanding, they also raise serious societal concerns \cite{stella2023using}. In particular, LLMs have been shown to exhibit harmful social biases, such as gender and racial biases, that reflect negative or stereotypical perceptions of certain groups \cite{garimella2021he}. These biases, which are inherited from the massive amounts of human corpora that LLMs are trained on \cite{bender2021dangers, dodge2021documenting, stella2023using, abramski2023cognitive}, mirror many of the social biases that humans possess \cite{acerbi2023large}, and they have the potential to reinforce the harmful narratives that are at the core of many social inequalities.

Several mitigation strategies have been developed for reducing biases in LLMs \cite{shen2023large}, but they have mostly focused on addressing explicit biases \cite{lin2025implicit, zhao2025explicit}, which are overt and easily detectable. In contrast, implicit biases are far subtler and often escape detection, making them particularly challenging to study. In humans, implicit biases reflect our subconscious associations and deep-rooted beliefs about the world, which often conflict with our explicit beliefs \cite{wilson2000model, perugini2005predictive}. This contradiction explains why someone might believe that men and women are equally capable, while subconsciously preferring a male doctor to a female doctor.
Similar to humans, implicit biases in LLMs can be nearly imperceptible \cite{stella2023using}, but they can manifest in subtle ways in downstream applications like question-answering, text-generation, and decision-making, even in models that exhibit little to no explicit bias \cite{lin2025implicit, bai2025explicitly, abramski2023cognitive, cau2025selective}. Moreover, since there is evidence that implicit biases influence behavior more than explicit biases \cite{greenwald2009understanding, stanley2011implicit, zhao2025explicit}, the cost of overlooking them can have serious consequences. This is especially a concern when LLMs are used in domains such as healthcare \cite{poulain2024bias}, recruiting \cite{anzenberg2025evaluating}, criminal justice \cite{struniawski2025bias}, psychology \cite{binz2025foundation}, and education \cite{lee2024life}, where social identity plays a key role. In such contexts, the application of biased LLMs can lead to harmful discriminatory outcomes, such as AI systems suggesting lower salaries for women and minorities \cite{sorokovikova2025surface}.

Another troubling and somewhat ironic consequence of LLM social biases has to do with the long-lasting impact they can have on our own biases. Recent studies have shown that as humans interact with biased LLMs, we subconsciously absorb the biases in their outputs, even long after we stop interacting with the models \cite{vicente2023humans}. These findings suggest the emergence of a feedback loop, in which LLMs acquire biases from humans, and through repeated exposure, we absorb new biases from LLMs. This dynamic is concerning, as it could potentially lead to a vicious cycle of reinforcing harmful social biases and even the emergence of new ones, especially if LLM biases are stronger than those possessed by humans. LLMs have also been shown to be more overconfident than humans when they are unsure about their answers, thus adding to the potential negative impacts that their biases could have on humans \cite{sun2025large}. These findings underscore the need to study LLM biases in relation to human biases, rather than evaluating models solely against neutral or artificial benchmarks.

The numerous harmful consequences of social biases in LLMs have led to an explosion of research dedicated to their detection and evaluation. In the last decade, great progress has been made towards this goal, but at the rate that LLMs are evolving, developing methodologies that remain relevant and effective is a continuous challenge. Because of the diversity of models, the growing number of their applications, and the various types of social biases that LLMs can exhibit, there is no one-size-fits all solution. Recent surveys on LLM bias evaluation methodologies emphasize the importance of a multifaceted approach that includes a wide range of evaluation methodologies that investigate bias from different perspectives \cite{guo2024bias, gallegos2024bias}. Such a multifaceted approach enables a broader more comprehensive understanding of the problem, which is crucial for informing effective bias mitigation strategies \cite{guo2024bias, gallegos2024bias}.

The word association network methodology that we present in this paper is a novel contribution to this aim. Our cognitive psychology-inspired methodology is an unexplored approach to evaluating implicit biases in LLMs that uses network models built from LLM-generated word associations. Our approach combines several elements from existing methodologies, but our method is unique in that it provides a metric that can be compared directly to humans, enabling parallel investigations between LLM and human cognition. Moreover, our methodology is flexible -- it can be used to investigate many different types of biases (e.g. gender, ethnicity, religion, politics, etc.) -- as well as scalable -- it can be applied to nearly any LLM, regardless of the application. In the remainder of this section, we provide additional background and motivation for our methodology. We first discuss existing work in the field of LLM bias evaluation to provide a contextual understanding of how our methodology compares to existing approaches. We then discuss the theories and methodologies from cognitive psychology upon which our methodology is based.

\subsection*{Overview of existing work on LLM bias evaluation}
Given the breadth and variety of research on this topic, the landscape of LLM bias evaluation methods can be difficult to navigate. Methodologies can be classified in many different ways based on the types of biases investigated and the data used to investigate them. An extensive review of bias evaluation methodologies is beyond the scope of this paper, and we refer interested readers to surveys on this topic \cite{gallegos2024bias, guo2024bias}. Rather, for the purpose of providing a contextual understanding of our word association network methodology, we discuss existing approaches to LLM bias evaluation as they relate to four defining characteristics of our methodology. Specifically, our methodology:
\begin{enumerate}
    \item evaluates \textit{intrinsic bias} as opposed to extrinsic bias
    \item uses an \textit{output-level} approach rather than a model-level approach
    \item evaluates bias via \textit{word associations}
    \item enables direct \textit{comparisons with humans}
\end{enumerate}

\noindent While none of these characteristics are unique to our methodology, the combination of them contributes to what distinguishes our approach from others. In what follows, we discuss the significance of these characteristics, focusing on the strengths and weaknesses of some of the methodologies that share them.

\subsubsection*{Intrinsic vs. extrinsic bias}
Methods for evaluating bias can be distinguished by the type of bias that they aim to investigate. In LLMs, biases can be classified as either intrinsic or extrinsic \cite{guo2024bias}. Intrinsic biases are those that are embedded within the model's architecture, acquired from the training data. They represent the patterns and relationships that are encoded in the internal representations of a model, and they are independent of any downstream task \cite{guo2024bias, lum2402bias}. Extrinsic biases are those that appear in applications and downstream tasks after fine-tuning; they can cause harm through biased outputs or decisions, and they manifest differently based on the downstream task \cite{guo2024bias, lum2402bias}. Thus, intrinsic bias evaluation methods typically aim to investigate the model's internal representations directly via their embedding space \cite{caliskan2017semantics, guo2021detecting, nadeem2020stereoset, kurita2019measuring}, while extrinsic bias evaluation methods aim to investigate biases that appear in specific tasks and contexts like question answering and text generation \cite{wan2023kelly, parrish2021bbq, zhao2018gender}. Although this distinction may resemble that between implicit and explicit bias, they are conceptually different. The intrinsic/extrinsic distinction concerns where the bias resides -- within model representations or outputs -- whereas the implicit/explicit distinction concerns how covertly or overtly bias is expressed. Both implicit and explicit biases can manifest in LLM outputs, thus, as extrinsic bias. However, implicit bias may be more closely aligned with intrinsic bias, since a model's representational structures represent latent associations, which form the basis of implicit bias. Recent works investigating the relationship between intrinsic and extrinsic evaluation metrics have found that the two measures do not always correlate, sparking skepticism about whether measures of intrinsic bias are really worthwhile \cite{cao2022intrinsic, steed2022upstream, kato2024analyzing, goldfarb2020intrinsic}. These findings suggest that evaluations of intrinsic bias are not sufficient for understanding how biases will manifest in outputs, and that extrinsic bias metrics may be better at predicting how biases can cause harm in specific tasks \cite{cao2022intrinsic, steed2022upstream}. However, these studies focused on comparisons involving only \textit{model-level} approaches, which access the model's representations directly via their embedding space. In contrast, our \textit{output-level} approach to intrinsic bias evaluation offers an alternative which creates proxies for the model's internal representations through careful prompting \cite{bai2025explicitly,abramski2023cognitive,abramski2024llm}, mirroring the behavioral methodology that is used for investigating implicit biases in humans. In what follows, we dive into the distinction between model-level and output-level approaches to provide a better understanding of how they are typically applied for investigating intrinsic and extrinsic bias, respectively.

\subsubsection*{Model-level vs. output-level approaches}
Methods for evaluating bias in LLMs can also differ by whether they operate at the model level, directly probing internal representations, or at the output level, treating the model as a black box and using outputs from prompting for investigating bias. Model-level methods are typically intrinsic since they directly access a model’s representational structures, while output-level approaches are typically extrinsic, focusing on specific downstream tasks or specific types of bias in certain domains (e.g. gender bias in recruiting) \cite{an2024large}. Model-level methods usually fall into one of two categories: embedding-based or probability-based \cite{gallegos2024bias}. Embedding-based metrics such as the well-known Word Embedding Association Test (WEAT) \cite{caliskan2017semantics} and similar variations \cite{may2019measuring, guo2021detecting} apply the logic of the Implicit Association Test (IAT) \cite{greenwald1998measuring}, a well-known metric for assessing implicit bias in humans. They measure vector similarities between social categories (e.g. gender) and attributes (e.g. careers) in different types of word embeddings. In contrast, probability-based metrics\cite{kurita2019measuring} use model-assigned probabilities to estimate biases. For example, masked token metrics ask a masked language model to complete fill-in-the-blank sentences \cite{kurita2019measuring, webster2020measuring}, and pseudo-log-likelihood metrics score the probability of generating a token given other words in the sentence \cite{nadeem2020stereoset, salazar2019masked}. Model-level approaches are advantageous because they offer fine-grained insights into how associations are structurally encoded in a model’s representations. However, they are increasingly difficult to apply to larger models where internal embeddings and probability distributions are not accessible, limiting their effectiveness at evaluating modern LLMs. Output-level methods, on the other hand, are not constrained by this lack of architectural transparency because they rely on LLM outputs generated through prompting rather than internal representations, making them more scalable and broadly applicable. This approach typically involves probing models with prompts that are designed to elicit biased or stereotypical outputs, and then analyzing the generated text \cite{gehman2020realtoxicityprompts, dhamala2021bold}. Common strategies include counterfactual testing, where demographic features in prompts are swapped to reveal unequal treatment \cite{zhou2024causalbench, banerjee2024all} and stereotype detection, which probes whether models reproduce or amplify social stereotypes \cite{bai2024fairmonitor, babonnaud2024bias}. Other approaches focus on sentiment and toxicity analysis \cite{inan2024llama, koh2024can} and acceptance/rejection rates, which measure disparities in decision-like tasks such as hiring or moral dilemmas \cite{an2024large, scherrer2023evaluating}. Since output-level methods are usually tailored for certain types of biases in specific contexts, they may be better at predicting real-world harm. However, this also makes them inflexible because they do not provide an overall measure of bias, and they cannot be easily extended to other less common forms of bias. Moreover, they tend to capture explicit rather than implicit biases, as they analyze surface-level outputs instead of latent semantic structures. Our method addresses these weaknesses by combining the scalability of output-level methods with the representational focus of intrinsic evaluation of implicit bias using word associations. In what follows, we discuss why word association methods are advantageous for studying implicit biases specifically.

\subsubsection*{Word association methods}
Methods of bias evaluation that use word associations are based on the Implicit Association Test (IAT), a behavioral methodology for evaluating implicit biases in humans \cite{greenwald1998measuring}. Built on the principle that shorter human reaction times correspond to semantic proximity between words, the test quantifies the strength of our unconscious associations between pairs of concepts, for example \textit{black} and \textit{bad} vs. \textit{white} and \textit{bad}. This methodology makes it possible to uncover implicit biases that lie below our level of consciousness, in contrast to explicit biases which are assessed by asking participants direct questions about their opinions and beliefs \cite{greenwald2009understanding}. Thus, applying the same logic as the IAT, methods that use word associations to evaluate biases in LLM are advantageous because they specifically target implicit rather than explicit bias, and they closely resemble the way that implicit biases are measured in humans, which is an important consideration given that LLMs behave in increasingly human-like ways \cite{bai2025explicitly}. Crucially, the IAT is a top-down approach, treating the mind as a black box and thus relying exclusively on behavior to make assumptions about the content of our conceptual representations. In contrast, model-level embedding-based evaluation metrics, like the WEAT \cite{caliskan2017semantics} use word associations, but they use a bottom-up approach, accessing semantic representations directly via the embedding space. As we discussed earlier, this bottom-up approach is problematic for modern LLMs with billions of parameters because the embeddings are not easily accessible. Instead, more recent output-level word association methods are a promising alternative because they use a top-down approach like the IAT, relying on outputs from prompting, which is akin to assessing behavior in humans. Output-level word association methods can take several forms. Some ask LLMs to fill in the blanks of masked sentence templates, for example, the analogy \textit{[X] are to nurses as [Y] are to surgeons} \cite{zhao2024comparative}. Others use ambiguous sentences like \textit{The boss called the employee because she got lost}, and then ask the model \textit{Who got lost?} to assess whether the model can recognize the sentence's ambiguity \cite{kotek2023gender}. Another method, called the LLM Word Association Test \cite{bai2025explicitly}, provides LLMs with a list of words (e.g. \textit{home, career, art, science}, etc.) and asks the model to choose which words to relate to it given two options (e.g. \textit{Julia} or \textit{Ben}) in order to assess a certain type of bias. It has been shown that these types of output-level measures of intrinsic bias are better at predicting real-world bias compared to model-level measures \cite{bai2025explicitly}. Word association methods for evaluating bias in LLMs is a promising direction of research, especially considering the parallels to the human way of assessing biases. However, what is still missing from these approaches is a metric that can be directly compared to humans. We discuss why comparisons with humans are important in the next section.

\subsubsection*{Comparisons with humans}
The importance of comparing LLM biases to human biases is widely recognized \cite{bas2024assessing, geva2025llms, tjuatja2024llms}. However, despite this recognition, work in this area tends to claim that LLMs demonstrate "human-like biases" without providing any concrete comparative metrics between humans and LLMs. Such quantitative and qualitative metrics are needed if we really want to understand the practical implications of interacting with biased LLMs, that is, understand if LLM biases are stronger, weaker, or categorically different than our own. Caliskan et al. (2017) \cite{caliskan2017semantics} were among the first to compare human biases to LLM biases, comparing the results of their WEAT methodology to human IAT results. However, Bai et al. (2025)\cite{bai2025explicitly}, who developed the LLM Word Association Test, have cautioned against making direct comparisons between LLM metrics and human IAT metrics given that the way the biases are measured differs significantly and it is not clear how to compare them \cite{bai2025explicitly}. Furthermore, they comment that future research connecting LLM biases to human biases could inform the design of bias intervention approaches in both LLMs and humans. Some recent work in this direction has begun to explore more direct comparisons between humans and LLMs. One study investigated differences between humans and LLMs in the level of confidence in their own answers, finding that LLMs are more overconfident than humans when their answers are incorrect \cite{sun2025large}. Additionally, there is a growing body of work that focuses on investigating differences in the cognitive capabilities between humans and LLMs using a top-down approach grounded in psychology. Work in this area has focused on investigating differences between humans and LLMs regarding the structure of the mental lexicon \cite{xiao2025human, abramski2024llm, abramski2025llm}, concreteness effects \cite{abramski2024llm}, and moral values \cite{xiang2025comparing}. However, this type of psychology-inspired comparative approach has yet to be applied for the specific purpose of evaluating implicit biases. The methodology we propose in this paper fills that gap. It uses a psychology-inspired methodology based on word associations to evaluate biases in such a way that allows direct quantitative and qualitative comparisons between humans and LLM.

\subsubsection*{Summary}
Up to this point, we have discussed existing methods for bias evaluation as they relate to the defining characteristics of the present methodology. To summarize, our methodology uses an output-level approach for intrinsic bias investigation, avoiding the problem of limited access to model representations and enabling the application to any LLM in various contexts and domains. Moreover, it uses word associations, so it is especially useful for investigating implicit rather than explicit bias, and it most closely resembles the IAT for assessing implicit biases in humans. And finally, it applies a psychology-inspired approach, allowing for direct quantitative and qualitative comparisons between humans and LLMs, unlike most approaches which offer only vague comparisons. Now that we have outlined the main characteristics of our methodology in relation to previous work, we lay the theoretical groundwork for our methodology in what follows.

\subsection*{Word association networks as models of implicit knowledge}

The methodology we propose involves measuring the strength of association between words by modeling semantic priming within LLM-generated word association networks. One of the strengths of our approach is that it is grounded in well-studied theories from cognitive psychology. Among them is the idea that conceptual knowledge is the foundation of language and thought \cite{fodor1998concepts,vigliocco2007semantic}. It happens that language models are also built on this idea -- language models acquire conceptual knowledge (i.e. semantic representations) from vast amounts of text, and then use that knowledge to generate linguistic output. Thus, models of conceptual knowledge can be used for studying cognitive and linguistic phenomenon in both humans and LLMs. Network models of semantic memory are one of the most widely-studied models of conceptual knowledge and have been applied extensively in cognitive science and AI \cite{collins1969retrieval}. In these network models, nodes represent concepts and edges represent relationships between them, so that concepts are defined in terms of how they relate to other concepts. Word association networks are a specific type of network model of semantic memory in which concepts are connected based on associative relations, like \textit{cat -- dog} (not to be confused with co-occurrence networks, in which concepts are connected based on how words co-occur in text). Word association networks are often built from free association norms, which are acquired by asking an agent -- human or machine -- to produce responses associated with a cue word. Since free associations have been shown to correlate with stable implicit attitudes \cite{schnabel2013free}, word association networks built from free associations can effectively model implicit knowledge. This makes them useful models for studying certain aspects of cognition that originate in implicit knowledge structures, such as implicit biases and stereotypes.

Within such network models of implicit knowledge, cognitive processes can be studied using the spreading activation theory of semantic processing \cite{collins1969retrieval, collins1975spreading}. According to the theory, when we are exposed to a concept, for example, by hearing a word, it leads to the activation of the concept's corresponding node in our semantic memory. The activation then spreads along the connections within our semantic memory network, decaying over time and activating other concepts along the way. The spreading activation theory explains the well-studied cognitive phenomenon of semantic priming, which is central to our methodology due to its connection to semantic relatedness. Semantic priming is our tendency to recognize certain concepts more quickly if we are first "primed" with a related concept \cite{mcnamara2005semantic, neely2012semantic}. For example, if you are primed with the concept \textit{hospital} and then asked to think of a profession, you are probably more likely to think of the concept \textit{doctor} than if you had been primed with the concept \textit{school}, because \textit{hospital} and  \textit{doctor} are more closely related.

The phenomenon of semantic priming has been widely studied using a behavioral experiment called the lexical decision task (LDT) \cite{hutchison2013semantic}, which measures the time it takes (reaction time) for participants to identify target words after being primed with a related or unrelated prime word, for thousands of prime-target pairs. However, more recently, a computational approach to studying semantic priming has been adopted by implementing a spreading activation process within network models of semantic memory \cite{siew2019spreadr, abramski2025llm}. The approach involves activating a prime node within a word association network and then observing the final activation level of the target node at the end of the spreading activation process. This framework effectively models semantic priming by simulating a lexical search process within a network model of semantic memory. Using this computational approach, the final activation levels of target nodes are greater, on average, when the corresponding prime node is related rather than unrelated to the target \cite{siew2019spreadr}. For example, we would expect the final activation level of the target node \textit{doctor} to be greater after activating the related prime node \textit{hospital} relative to the unrelated prime node \textit{school}. Crucially, this computational approach has been empirically validated by correlating final activation levels from the computational approach with reaction times from the LDT behavioral experiment \cite{hutchison2013semantic}, for the same prime-target pairs. This correlation was first found by modeling semantic priming within human-generated word association networks \cite{siew2019spreadr, nelson2004university}, and it was later extended to LLM-generated word association networks \cite{abramski2025llm}. This grounding in behavioral data is fundamental because it ensures that the computational approach accurately represents the real cognitive phenomenon that it aims to model.

\begin{figure}[H]
\centering
\includegraphics[width=.7\linewidth]{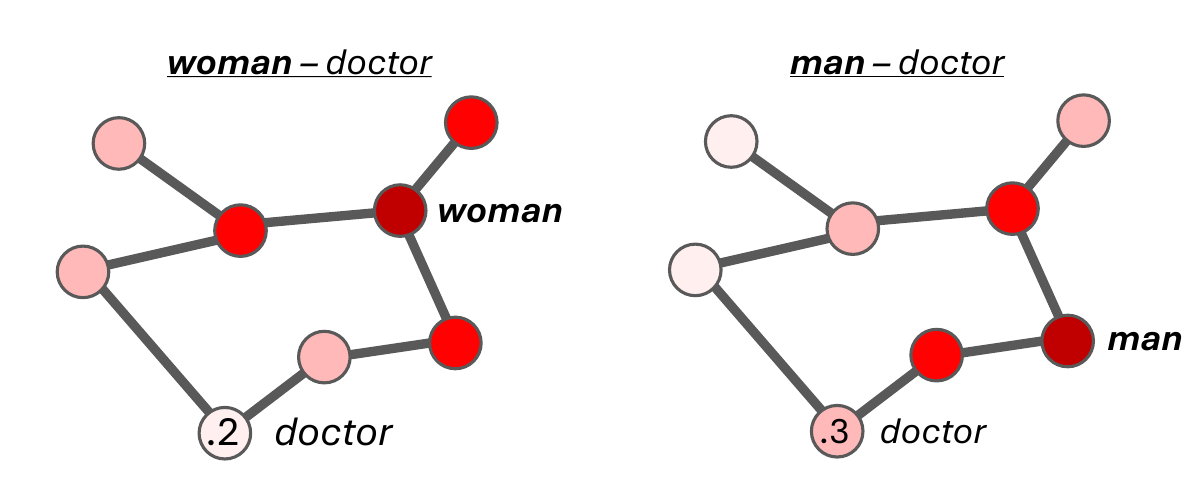}
\caption{\textbf{Simplified example of gender bias evaluation.} The simplified example shows how our methodology can be used to evaluate gender bias by modeling semantic priming within a word association network. A spreading activation process is implemented in the network by activating the prime nodes \textit{woman} and \textit{man} independently. At the end of the process, the final activation level of \textit{doctor} is 0.2  when \textit{woman} is the prime node,  compared to 0.3 when \textit{man} is the prime node. This shows that the association between \textit{man} and \textit{doctor} is stronger than the association between \textit{woman} and \textit{doctor}, indicating gender bias.}
\label{fig:spreading}
\end{figure}

For the purposes of our methodology, we adopt this computational approach that models semantic priming by implementing spreading activation within word association networks, but we adapt it for evaluating implicit biases. Since semantic relatedness is the cognitive mechanism that drives semantic priming, then the final activation levels of target nodes can be considered a proxy for semantic relatedness. Thus, the final activation levels can be considered a measure of the strength of association between prime-target pairs. By choosing prime-target pairs that relate to certain social biases, for example \textit{man -- doctor} and \textit{woman -- doctor}, we can quantify implicit biases by observing relative differences in the activation levels of the prime-target pairs, as shown in the example in Figure \ref{fig:spreading}.

This approach follows the same principle as the IAT that we discussed earlier, but since our methodology is computational rather than behavioral in nature, it enables more flexibility, scalability, and reproducibility. It also makes it possible to apply the approach to both LLMs and humans, by using word association networks built from free associations generated by both LLMs and humans. Additionally, since the word association networks used in this methodology are models of broad implicit knowledge including not only the primes and targets of interest, but thousands of other concepts, we can go beyond simple measures of word associations by examining the conceptual paths that are traveled in order to get from primes to targets. Thus, in addition to the quantitative metrics of bias that our methodology provides, it also provides a qualitative understanding of the nature of implicit biases, which is a key feature of our methodology that other approaches lack.

In the remainder of this paper, we demonstrate the utility of our comparative methodology by applying it to three LLMs (Mistral, Llama3, and Haiku) as well as humans to investigate social biases related to gender, religion, ethnicity, sexual orientation, and political party.
Thus, the remainder of this paper is organized as follows. In the Methods section, we describe in detail the three-step process that our methodology entails, including 1) network building, 2) spreading activation, and 3) bias evaluation. In the Results section, we present the results obtained from applying our comparative methodology to the three LLMs and humans for evaluating implicit social biases. Finally, in the Discussion section, we discuss limitations, future work, and the impact that our methodology could have on ensuring the ethical and trustworthy development of LLMs.

\section{Methods} \label{sec:methods}

In this section, we describe how we apply our methodology to the three LLMs (Mistral, Llama3, and Haiku) and humans to comparatively evaluate their implicit biases related to gender, religion, ethnicity, sexual orientation, and political party. We apply three different versions of our methodology: one approach that evaluates bias in the form of stereotypes, another that evaluates bias in the form of positive or negative perceptions, otherwise known as valence, and a third that evaluates bias in the form of emotions. The stereotypes approach and emotions approach are better suited when the social identity of interest has only two categories, while the valence approach works well when the social identity of interest has two or more categories. Thus, to demonstrate our methodology, we apply the stereotypes approach for evaluating gender biases, the valence approach for evaluating biases related to religion, ethnicity, and sexual orientation, and the emotions approach for evaluating biases related to political party. The three different approaches of our methodology provide a wider view of implicit biases by enabling investigations of different ways in which biases can be expressed: through stereotypes, valence, or emotion. They also make our methodology flexible and adaptable, enabling applications to social identities with a varying number of categories.

\begin{figure}[H]
\centering
\includegraphics[width=1\linewidth]{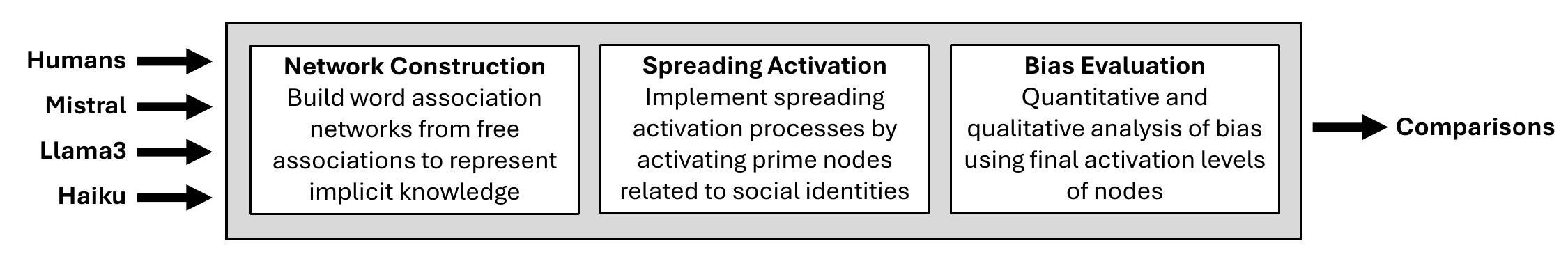}
\caption{\textbf{Methodological overview}. The flowchart shows an overview of the three-step methodology for evaluating biases in LLMs compared to humans. For humans and each LLM (Mistral, Llama3, and Haiku), first, word association networks are built from free association norms, then spreading activation processes are implemented within the networks, and finally the final activation levels of nodes in the networks are analyzed to evaluate biases. This evaluation methodology enables direct comparisons of social biases between the various LLMs and humans.}
\label{fig:methods}
\end{figure}

Our methodology can be broken down into three steps -- network construction, spreading activation, and bias evaluation -- as shown in Figure \ref{fig:methods}. The first step entails building word association networks from free association norms. We use existing datasets of human-generated \cite{de2019small} and LLM-generated \cite{abramski2025llm} free associations that were generated following identical methodologies, making them directly comparable. These network models represent the implicit knowledge of concepts possessed by humans and LLMs, respectively. The second step entails implementing spreading activation processes by activating prime nodes within the networks to model semantic priming. The prime nodes correspond to different social categories of each of the social identities (i.e. gender, religion, ethnicity, sexual orientation, and political party), and they are shown in \ref{tab:primes}. For each network, this step yields final activation levels for each node in the network for each activated prime, stored in an activation level matrix. These values of the matrix represent the strength of association between the activated prime nodes and all other nodes in the network. Finally, the third step entails evaluating the bias by conducting statistical analyses of the activation level matrices to produce quantitative bias evaluation metrics. Only at this step does the methodology diverge for the three different approaches (stereotypes, valence, and emotions) because the different approaches require different types of statistical analyses of the matrices. This step also includes a qualitative analysis of bias for the stereotypes approach. 
By applying this three-step methodology to humans and each of the three LLMs, the resulting bias evaluation metrics enable direct comparisons between each of the LLMs and humans. In what follows, we dive into each step of the three-step methodology.

\subsection*{Network construction}
The first step of the methodology is network construction, which entails building word association networks from free association norms. In these networks, nodes represent concepts, and edges represent associative relations between them. 
The networks are built from existing datasets of free association norms generated by humans and LLMs, respectively.

To build the human network, the Small World of Worlds (SWOW)\cite{de2019small} dataset of human-generated free association norms is used, while for the LLM networks, the LLM World of Words (LWOW)\cite{abramski2025llm} dataset of LLM-generated free association norms is used. The LWOW dataset was derived from the SWOW dataset with the purpose of enabling comparative analyses between humans and LLMs. The SWOW dataset was created through a behavioral experiment in which humans were provided with a cue word and asked to respond with three words that come to mind in response to the cue word, thus tapping into the implicit conceptual knowledge possessed by participants. This experiment was repeated for roughly 12,000 unique cue words, 100 times for each cue word, yielding a dataset of roughly 3 million responses. The LWOW dataset was created by simulating the same behavioral experiment used to create the SWOW dataset, including the same cue words and the same number of repetitions, but instead of using human participants, three different LLMs (Mistral, Llama3, and Haiku) were prompted to provide responses to the cue words, thus tapping into the implicit semantic representations encoded in the LLM architecture. The creation of the LWOW entailed applying a preprocessing procedure to the data in order to facilitate network analyses (lemmatization, spelling corrections, etc.), and in order to align the SWOW dataset with the LWOW, this same preprocessing procedure was applied to the original SWOW. This resulted in the creation of a modified SWOW dataset that is directly comparable to the LWOW dataset. Thus, the free association norms used to create the networks for our methodology include a modified version of the SWOW dataset and the three LWOW datasets. The three LLM datasets can be downloaded from the LWOW github page \href{https://github.com/LLMWorldOfWords/LWOW}{https://github.com/LLMWorldOfWords/LWOW}. The modified SWOW dataset, while not available directly for download because of license restrictions, can be generated by following the procedure described on the LWOW github page.

From those four datasets of free association norms, we build word association networks for humans and all three LLMs by connecting cue words to their responses. Edges are weighted to reflect the frequency of each response to each cue, so if the word \textit{cat} appeared 20 times as the response to the cue \textit{dog}, then the edge from \textit{dog} to \textit{cat} would have a weight of 20. Edges are naturally directed from cues to responses, but in order to facilitate the spreading activation analyses in the next step of the methodology, directed edges are transformed into undirected edges, maintaining the largest of the two edge weights in the case of bidirectional edges. A filtering procedure is then applied to the networks in order to remove unwanted nodes and edges, and to reduce the size of the networks making them more computationally manageable. First, nonsensical and uncommon responses are filtered out by keeping only nodes in WordNet, a large lexical database of English words. Then, idiosyncratic edges (i.e. edges with weight = 1) are removed to ensure that the association is something shared among two or more people (or iterations in the case of the LLMs). Finally, we take the largest connected component of the network, which is essential for the spreading activation processes in the next step of the methodology. The network statistics of the resulting word association networks are shown in Table \ref{tab:net_stats}. 

\begin{table}[h]
\centering
\begin{tabular}{lcccc}
\hline
\textbf{Network} & \textbf{Nodes} & \textbf{Edges} & \textbf{Density} & \textbf{Average degree} \\ \hline
Humans           & 24,308         & 317,344        & 0.0011           & 26                      \\
Mistral          & 20,339         & 199,103        & 0.0010           & 20                      \\
Llama3           & 38,987         & 546,866        & 0.0007           & 28                      \\
Haiku            & 15,596         & 64,599         & 0.0005           & 8                       \\ \hline

\end{tabular}
\caption{\textbf{Network statistics.} The table shows network statistics for each of the word association networks, including the number of nodes and edges, the network density, and the average degree.}
\label{tab:net_stats}
\end{table}

\subsection*{Spreading activation}

The second step of the methodology involves implementing spreading activation processes within the word association networks by activating prime nodes related to the social identities of interest. As we discussed earlier, this spreading activation process is used to model semantic priming, and it represents what happens in our semantic memory when we are exposed to a concept, such as when we hear a word. Thus, by activating prime nodes that are related to certain social identities, we can measure how strongly certain concepts (i.e. target nodes) are associated with the prime nodes by observing the final activation levels of the target nodes at the end of the spreading activation process. This process is implemented as follows.

First, a set of prime nodes must be defined. The prime nodes should correspond to different categories of the social identity of interest. For example, \textit{european}, \textit{african}, and \textit{asian} are different categories of the social identity ethnicity. When considering only two categories of a social identity, such as is often the case with gender\footnote{For the purpose of demonstrating our methodology, we do not include non-binary gender identities that are outside of the male/female gender dichotomy.}, the prime nodes should be one or more pairs of opposing concepts related to the two categories, such as \textit{woman/man}, \textit{female/male}, etc. Rather, when considering more than two categories, the prime nodes should be a set of three or more categories. Crucially, the prime nodes should be of the same linguistic function, so for example, \textit{woman} should not be paired with \textit{sir}, but rather its antonym, \textit{man}, and \textit{buddhism} should not be included in the set containing \textit{christian}, \textit{muslim}, and \textit{jewish}, but rather \textit{buddhist} in order to match the linguistic function of the other prime nodes. This is important in order to limit confounding that could be introduced by differences in linguistic functions. It is also important to note that the prime nodes must be present in the networks in order for the methodology to be applied, so that should also be kept in mind when defining the prime nodes. To demonstrate our methodology, we investigate five different social identities: gender, religion, ethnicity, sexual orientation, and political party. The sets of prime nodes defined for these social identities, as well as the approaches used, are shown in Table \ref{tab:primes}. Note that for gender the prime nodes are five sets of opposing concepts related to female and male, while for religion, ethnicity, and sexual orientation, the primes nodes are sets of categories, and for political party, the prime nodes include a single pair of opposing concepts.

\begin{table}[H]
\centering
\begin{tabular}{lll}
\hline
\textbf{Approach} & \textbf{Social identity} & \textbf{Prime nodes}                                                                                                 \\ \hline
stereotypes        & gender                   & \begin{tabular}[c]{@{}l@{}}woman, man\\ female, male\\ girl, boy\\ mother, father\\ feminine, masculine\end{tabular} \\ \hline
\textbf{}         & religion                 & christian, muslim, buddhist, jewish, athiest                                                                         \\ \cline{2-3} 
valence           & ethnicity                & european, african, hispanic,   asian, indigenous                                                                     \\ \cline{2-3} 
                  & sexual orientation       & straight, gay, lesbian, bisexual                                                                                     \\ \hline
emotions          & political party          & democrat, republican                                                                                                 \\ \hline
\end{tabular}
\caption{\textbf{Prime nodes for different social identities.} The table outlines the three different versions of our methodology that are applied for evaluating biases related to five different social identities. The activated prime nodes are shown as well.}
\label{tab:primes}
\end{table}

Once the prime nodes are defined, they are activated individually and independently within the word association network. A prime node is activated within the network by assigning it a certain activation level. The activation then spreads throughout the network iteratively. At each iteration, some percentage of the activation is retained by the activated node, while the rest is passed to its neighboring nodes in a percentage that is proportional to the weights of its connections. This process continues iteratively for a finite number of time steps, and at the last time step, all nodes in the network converge on a final activation level. Computationally, we implemented the spreading activation processes in the word association networks using the R package \textit{spreadr}\cite{siew2019spreadr}, but the more recent python package \textit{SpreadPy}\cite{citraro2025spreadpy} could be used as well. The \textit{spreadr} package requires the specification of the parameters needed for the spreading activation process, including: the initial activation level assigned to the prime node, which we set to the number of nodes in the network; the number of time steps, which we set to twice the diameter of the network; and the percentage of activation retained by the activated node, which we set to the default of 0.5.

At the end of the spreading activation process for a given prime node, the final activation levels (ALs) of all nodes are stored as a vector that represents the strength of association between the prime node and each node in the network. This process is repeated for each prime node, resulting in a vector for each prime node. The vectors are combined into a final activation level matrix such that the rows are the nodes of the network, the columns are the prime nodes, and the entries of the matrix reflect the strength of association between the prime nodes and each node in the network. At this point, any node in the network can be considered a target node, and the relative strength of association between certain prime-target pairs can be used to evaluate implicit biases. We discuss how the activation level matrices are used to quantitatively and qualitatively evaluate implicit biases in the next section.

\subsection*{Bias evaluation}

The third step of the methodology involves conducting statistical analyses of the final activation level matrices to quantitatively and qualitatively evaluate implicit biases. The purpose of the statistical analyses is to identify significant patterns in the final ALs that indicate the presence of implicit biases towards certain social categories. The way this is done differs for the three different approaches.

\subsubsection*{Stereotypes approach}

We apply the stereotypes approach for evaluating biases related to gender. Stereotype bias involves associating certain groups with certain traits or concepts, for example, associating women with emotion and associating men with strength. For our methodology, that translates to defining prime nodes that relate to certain social identities and target nodes that relate to certain stereotypical traits or concepts. For the prime nodes, we define five matched pairs of female-related and male-related primes (e.g. \textit{woman/man}, \textit{female/male}) as  shown in Table \ref{tab:primes}. For the target nodes, we define 25 stereotypical female-related adjectives and 25 stereotypical male-related adjectives \cite{gaucher2011evidence}, shown in Table \ref{tab:gender_primes_tgts}. Considering all possible combinations of the 10 prime nodes and the 50 target nodes, we get a total of 500 prime-target pairs. Half of those pairs are stereotype-consistent, for example \textit{feminine -- compassionate} and \textit{masculine -- forceful}, while the other half are stereotype-inconsistent, for example \textit{feminine -- forceful} and \textit{masculine -- compassionate}. The aim of the stereotypes approach is to quantify differences in the final ALs of the stereotype-consistent prime-target pairs compared to the stereotype-inconsistent prime-target pairs. Higher final ALs of the stereotype-consistent prime-target pairs compared to the stereotype-inconsistent prime-target pairs is indicative of stereotype bias. The greater the discrepancy, the greater the stereotype bias.

We measure this stereotype bias using a quantitative metric obtained by computing paired differences in the final ALs of targets, and then computing the effect sizes of the differences. First, the final activation level matrix is normalized using the L2 norm by normalizing the columns then rows of the matrix. This normalization accounts for differences in the centrality of nodes within the word association networks, ensuring that the normalized final ALs reflect effects from semantic priming and not effects related to node centrality, like frequency effects. Then, paired differences in normalized final ALs are computed separately for all 25 female-related targets \( T_F \) and all 25 male-related targets \( T_M \) as follows:
\[
\text{Diff}_F(t) = \text{AL}_{p_F}(t) - \text{AL}_{p_M}(t)
\quad \forall \, t \in T_F, \; (p_F, p_M) \in P
\]
\[
\text{Diff}_M(t) = \text{AL}_{p_M}(t) - \text{AL}_{p_F}(t)
\quad \forall \, t \in T_M, \; (p_F, p_M) \in P
\]
where \( AL_{p}(t) \) denotes the final activation level of target \( t \) after being activated by prime \( p \). Differences are computed for all five matched pairs of female-related and male-related primes \( (p_F, p_M) \in P \), resulting in two sets of 125 paired differences. Positive differences are indicative of stereotype bias, since they represent greater strength of association between stereotype-consistent prime-target pairs. The Wilcoxon test for paired samples is applied to each set of differences and the effect sizes are computed. The resulting effect sizes, one for female-related targets and one for male-related targets, are quantitative metrics for female-related and male-related stereotype biases, respectively.

This quantitative metric provides a concrete and objective measure that enables direct comparisons between different LLMs and humans. However, we can gain a deeper understanding of the nature of the stereotype biases that we observe through a qualitative analysis of the paths traveled within the networks to get from a prime node to a target node. These prime-target paths reveal certain nuances about the implicit knowledge structures underlying stereotype biases. For example, by observing the paths from the prime node \textit{female} to the target node \textit{compassionate} and from the prime node \textit{male} to the target node \textit{compassionate}, we can gain an understanding of the contexts through which certain implicit associations between concepts are made. Together, the quantitative metric and the qualitative analysis of paths are a powerful tool for implicit stereotype bias evaluation. 

\begin{table}[H]
\centering
\begin{tabular}{ll}
\hline
                        & \multicolumn{1}{c}{\textbf{Target nodes}}                                                                                                                                                                                                                                                                              \\ \hline
\textbf{Female-related} & \begin{tabular}[c]{@{}l@{}}affectionate,  cheerful, compassionate, considerate, cooperative, emotional, empathetic,\\ gentle, honest, kind, loyal, modest, nagging, nurturing, pleasant, polite, quiet, sensitive,\\ submissive, supportive, sympathetic, tender, trusting,   understanding, warm,\end{tabular}        \\ \hline
\textbf{Male-related}   & \begin{tabular}[c]{@{}l@{}}active, aggressive, ambitious, analytical, assertive, athletic, competitive, confident,\\ courageous, decisive, determined, dominant, forceful, greedy, hostile, impulsive,\\ independent, intellectual, leader, logical, outspoken, persistent, reckless, stubborn, superior,\end{tabular} \\ \hline
\end{tabular}
\caption{\textbf{Gender-related stereotype adjectives}. The table shows stereotypical female-related and male-related adjectives that are used as target nodes in the stereotypes approach to evaluating gender biases. The adjectives come from a previous study investigating gendered language in job postings \cite{gaucher2011evidence}.}
\label{tab:gender_primes_tgts}
\end{table}

\subsubsection*{Valence approach}

We apply the valence approach for evaluating biases related to religion, ethnicity, and sexual orientation. Contrary to stereotype bias, biases of valence do not involve specific traits or concepts, like the target nodes defined in the previous analysis. Instead, valence has to do with general positive or negative perceptions of certain groups. For example, to measure biases of valence, we are interested in determining whether certain social categories, such as \textit{christian}, are generally associated with more positive concepts, while other social categories such as \textit{muslim} are generally associated with more negative concepts. Since we are not interested in associations related to a certain trait or concept, there is no need to identify specific target nodes. Rather, we are interested in how the final ALs of all the nodes vary with respect to their valence ratings after being activated by different prime nodes. The sets of prime nodes defined for the different categories of religion, ethnicity, and sexual orientation are shown in Table \ref{tab:primes}. We obtain valence ratings for nodes of the network from the VAD lexicon dataset \cite{mohammad2018obtaining}, in which words are rated by humans from 0 to 1 with 0 corresponding to negative valence and 1 corresponding to positive valence. Thus, using these ratings and the final ALs of all nodes, we can measure how valence differs across different categories of the social identities of interest.

To quantify this valence bias, we fit generalized linear models (GLMs) to measure the relationship between the final ALs of the network nodes and their valence ratings. First, as with the stereotypes approach, the final AL matrix is normalized using the L2 norm by normalizing the columns then rows of the matrix. Then, for each social identity (religion, ethnicity, and sexual orientation) we fit separate univariate GLMs for each social category (prime node), such that the response variable is the vector of valence scores of all nodes, and the predictor is the vector of final ALs of all nodes after being activated by a given prime node. Naturally, the nodes that do not have a valence rating are excluded from the GLM. The resulting coefficient value of each GLM indicates the relationship between the final ALs of the nodes and the valence ratings. Positive coefficients indicate that the corresponding social category (i.e \textit{christian}) is perceived positively, while negative coefficients indicate that it is perceived negatively (relative to the other social categories). Large differences in perceptions across the different social categories (some very positive coefficients and some very negative coefficients) indicate strong valence bias. The greater the variation in coefficients, the greater the bias. For each social identity, this bias is quantified by fitting a single multivariate GLM with all social categories (prime nodes) and then computing the effect size by conducting a Wald test, with the null hypothesis that all coefficients are equal. Thus, a large effect size tells us that the coefficients differ significantly from each other, and so the different social categories are perceived differently in terms of their valence. The resulting effect size is thus a measure of the bias of valence regarding the social identity of interest, enabling direct comparisons between different LLMs and humans.

\subsubsection*{Emotions approach}
We apply the emotions approach for evaluating biases related to political party, specifically, the two US political parties, Democrats and Republicans. This approach combines elements from both the stereotypes approach and the valence approach. The general idea is the same as the valence approach, but instead of evaluating positive and negative perceptions of certain groups, it evaluates feelings towards certain groups considering eight core emotions, defined by Plutchik\cite{plutchik1980general}: anger, anticipation, disgust, fear, joy, sadness, surprise, and trust. The computation of the quantitative metric is more similar to the stereotypes approach since it is applied to a social identity with only two categories. In this case, we have a single pair of prime nodes -- \textit{democrat} and \textit{republican} -- so the aim of this approach is to measure differences in emotions towards these two different categories. In order to do this, we identify sets of emotion-related target nodes, which are words in the networks that elicit certain emotions, according to human judgments from the  NRC emotion lexicon \cite{mohammad2018obtaining}. Thus, for each of the eight emotions, a set of nearly 1,000 emotion-related targets are defined. 

To quantify differences in emotions towards Democrats and Republicans, we compute differences in the final ALs of the emotion-related targets after being activated by \textit{democrat} compared to \textit{republican}. First, the final activation level matrix is normalized first by columns and then by rows, but instead of using the L2 norm, we found that the L1 norm was better suited for the emotions approach (we expand on this topic in the discussion). Then, for each emotion \( e \) in the set of eight emotions, paired differences in normalized final ALs are computed for all emotion-related targets \( T_e \)  as follows:
\[
\text{Diff}_e(t) = \text{AL}_{democrat}(t) - \text{AL}_{republican}(t)
\quad \forall \, t \in T_e \
\]
where \( AL_{p}(t) \) denotes the final activation level of target \( t \) after being activated by prime \( p \). This results in eight sets of nearly 1,000 differences, one set for each emotion, such that positive differences indicate stronger emotions towards Democrats while negative differences indicated stronger emotions towards Republicans. The Wilcoxon test for paired samples is applied to each set of differences and the effect sizes are computed. The effect size, which can be either positive or negative, is therefore a measure of how strongly an emotion is felt for either Democrats (positive effect) or Republicans (negative effect).

\section{Results}
\label{sec:results}

\subsubsection*{Stereotypes related to gender}

\begin{figure}[H]
\centering
\includegraphics[width=.6\linewidth]{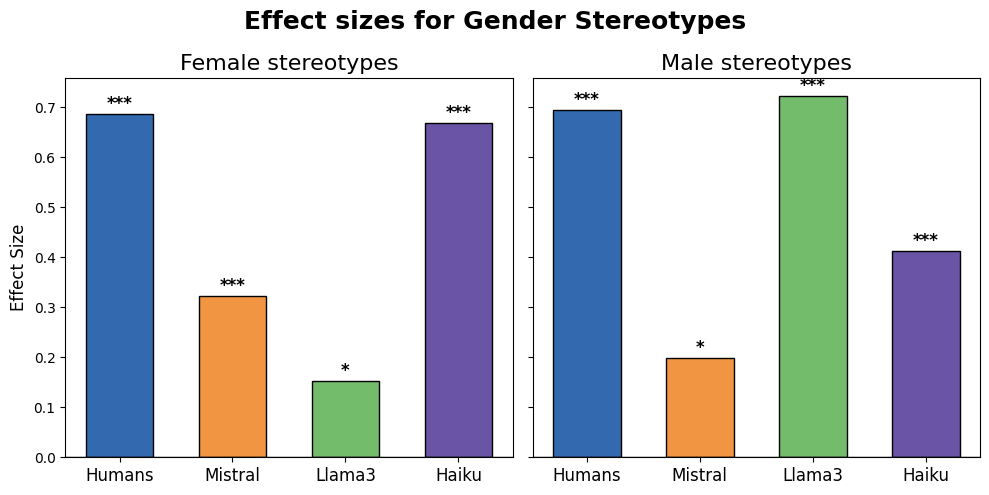}
\caption{\textbf{Effect sizes for gender stereotypes.} The barplots show the effect sizes for female-related stereotypes (left) and male-related stereotypes for humans and the three LLMs. Statistical significance is indicated with stars.}
\label{fig:effect_gender}
\end{figure}

We applied the stereotypes approach for evaluating gender bias in humans, Mistral, Llama3, and Haiku. We implemented spreading activation processes in word association networks generated by humans and the three LLMs by activating prime nodes related to gender in each of the four networks. We then observed the final ALs of female-related and male-related stereotypical adjectives. We quantified stereotype biases by computing differences in final ALs of stereotypical and non-stereotypical prime-target pairs. The Wilcoxon test for paired samples was applied to the differences and effect sizes were computed separately to quantify female-related stereotypes and male-related stereotypes. The effect sizes are shown in the barplots in Figure \ref{fig:effect_gender}. Positive significant effects indicating the presence of stereotype biases are observed in all four of the networks. Effect sizes for female stereotypes and male stereotypes are nearly identical for humans, but they vary in the LLMs. Female stereotypes are stronger than male stereotypes in Mistral and Haiku, but male stereotypes are stronger than female stereotypes in Llama3. Overall, humans demonstrate the strongest gender stereotypes while Mistral demonstrates the weakest gender stereotypes.

\begin{figure}[H]
\centering
\includegraphics[width=.8\linewidth]{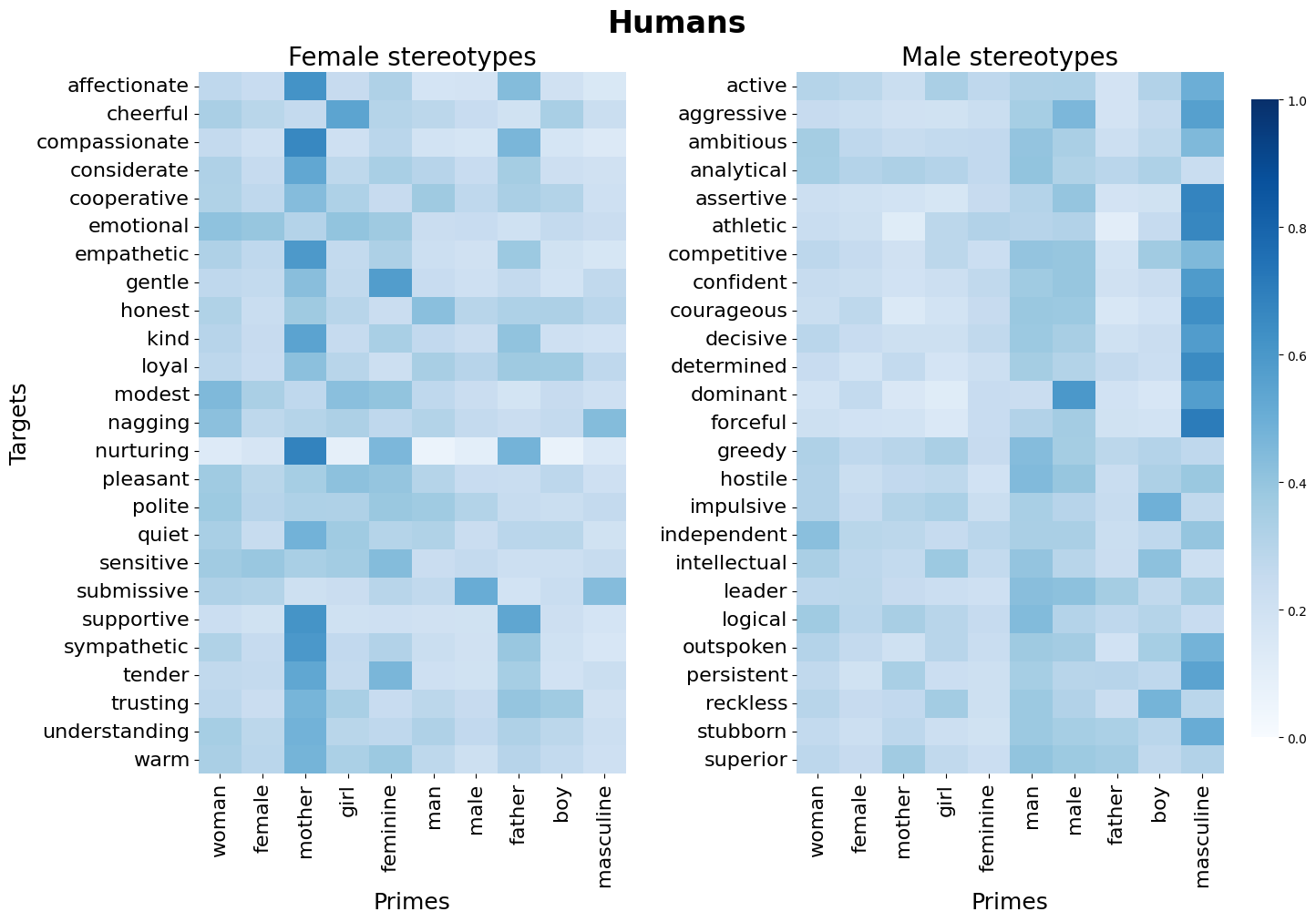}
\caption{\textbf{Gender biases in humans.} The heatmaps show the normalized final activation levels of 25 female-related targets (left) and 25 male-related targets (right) after activating 10 gender-related prime nodes in the human network.}
\label{fig:heat_humans}
\end{figure}

\begin{figure}[H]
\centering
\includegraphics[width=.8\linewidth]{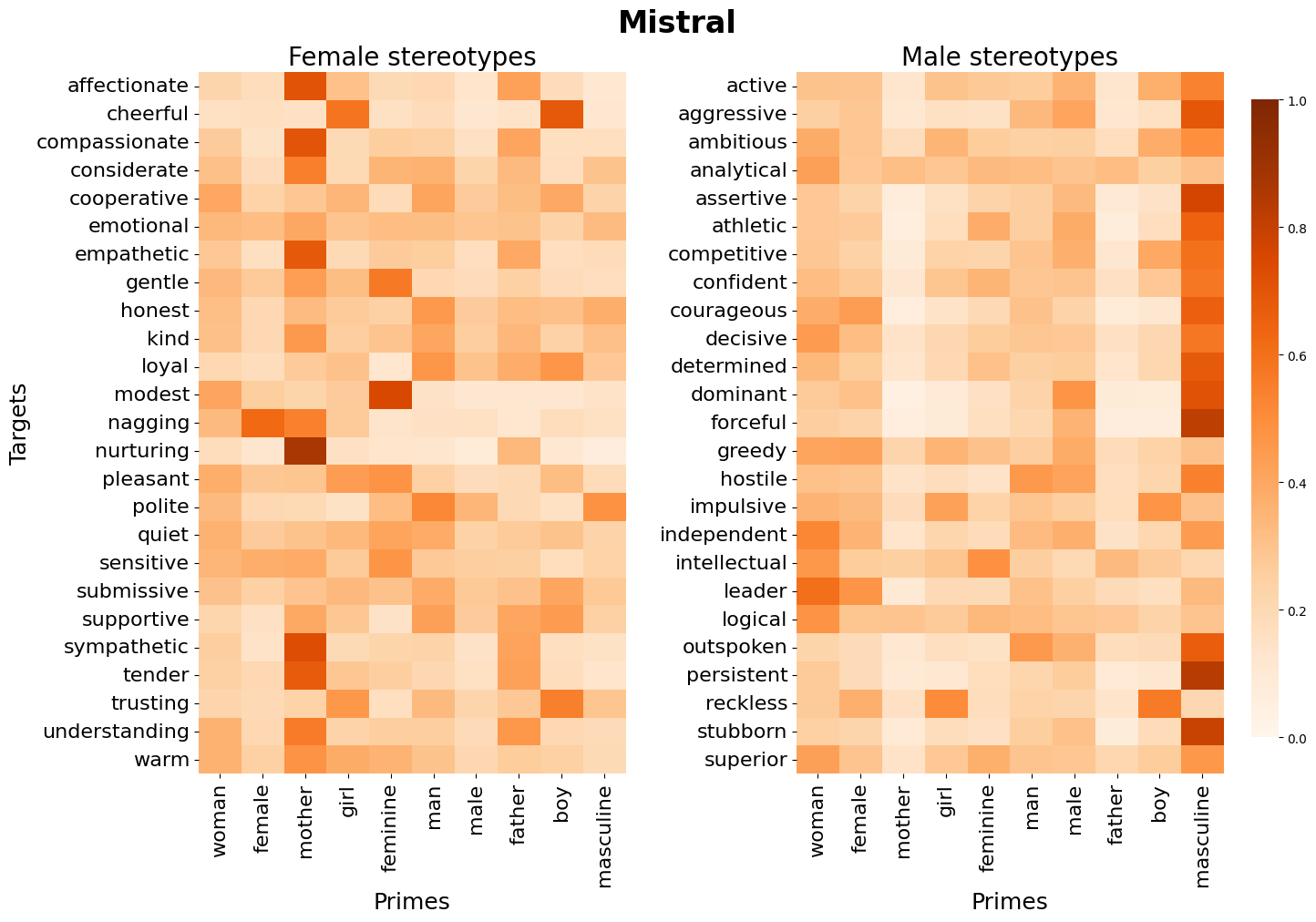}
\caption{\textbf{Gender biases in Mistral.} The heatmaps show the normalized final activation levels of 25 female-related targets (left) and 25 male-related targets (right) after activating 10 gender-related prime nodes in the Mistral network.}
\label{fig:heat_mistral}
\end{figure}

\begin{figure}[H]
\centering
\includegraphics[width=.8\linewidth]{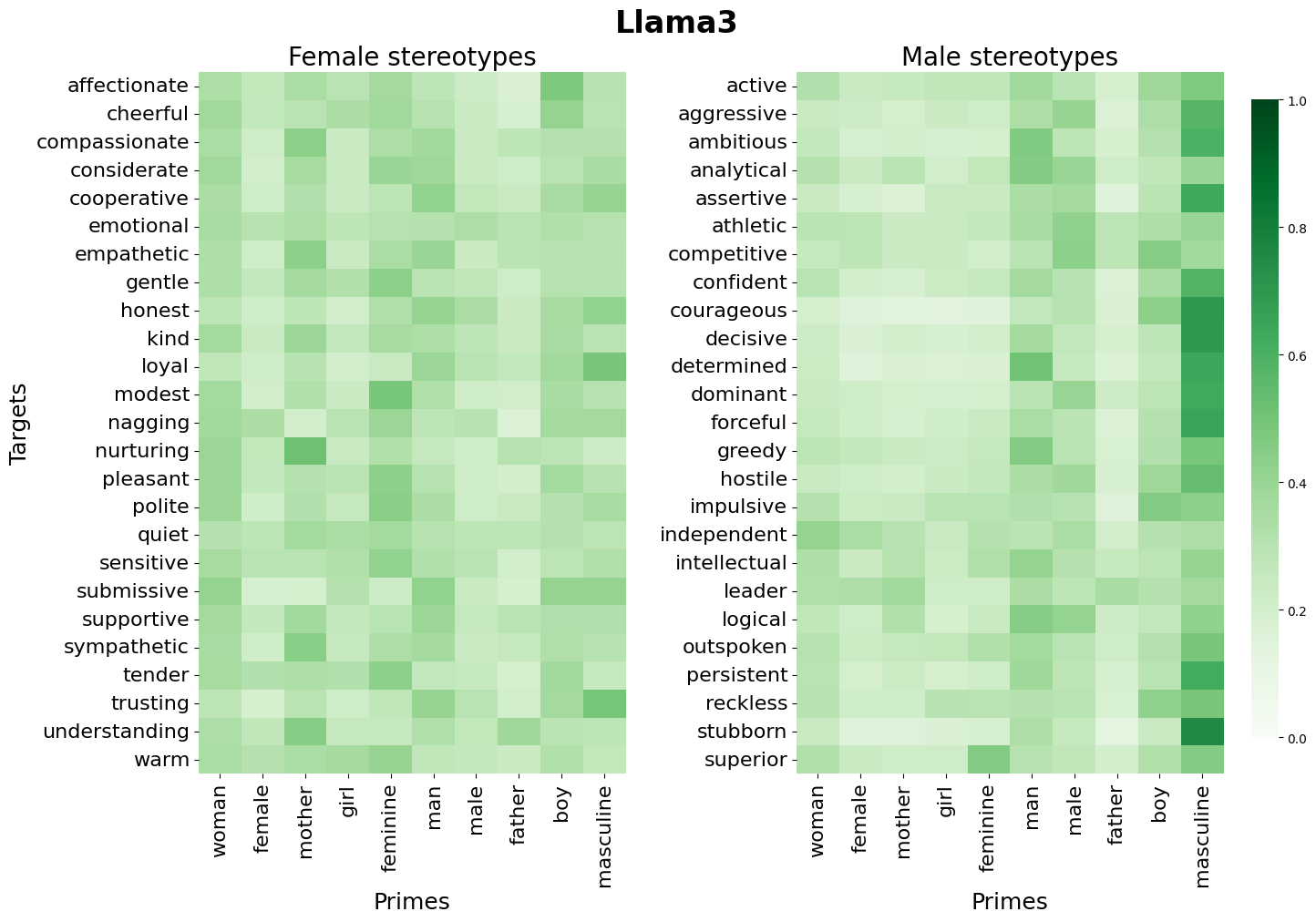}
\caption{\textbf{Gender biases in Llama3.} The heatmaps show the normalized final activation levels of 25 female-related targets (left) and 25 male-related targets (right) after activating 10 gender-related prime nodes in the Llama3 network.}
\label{fig:heat_llama3}
\end{figure}

\begin{figure}[H]
\centering
\includegraphics[width=.8\linewidth]{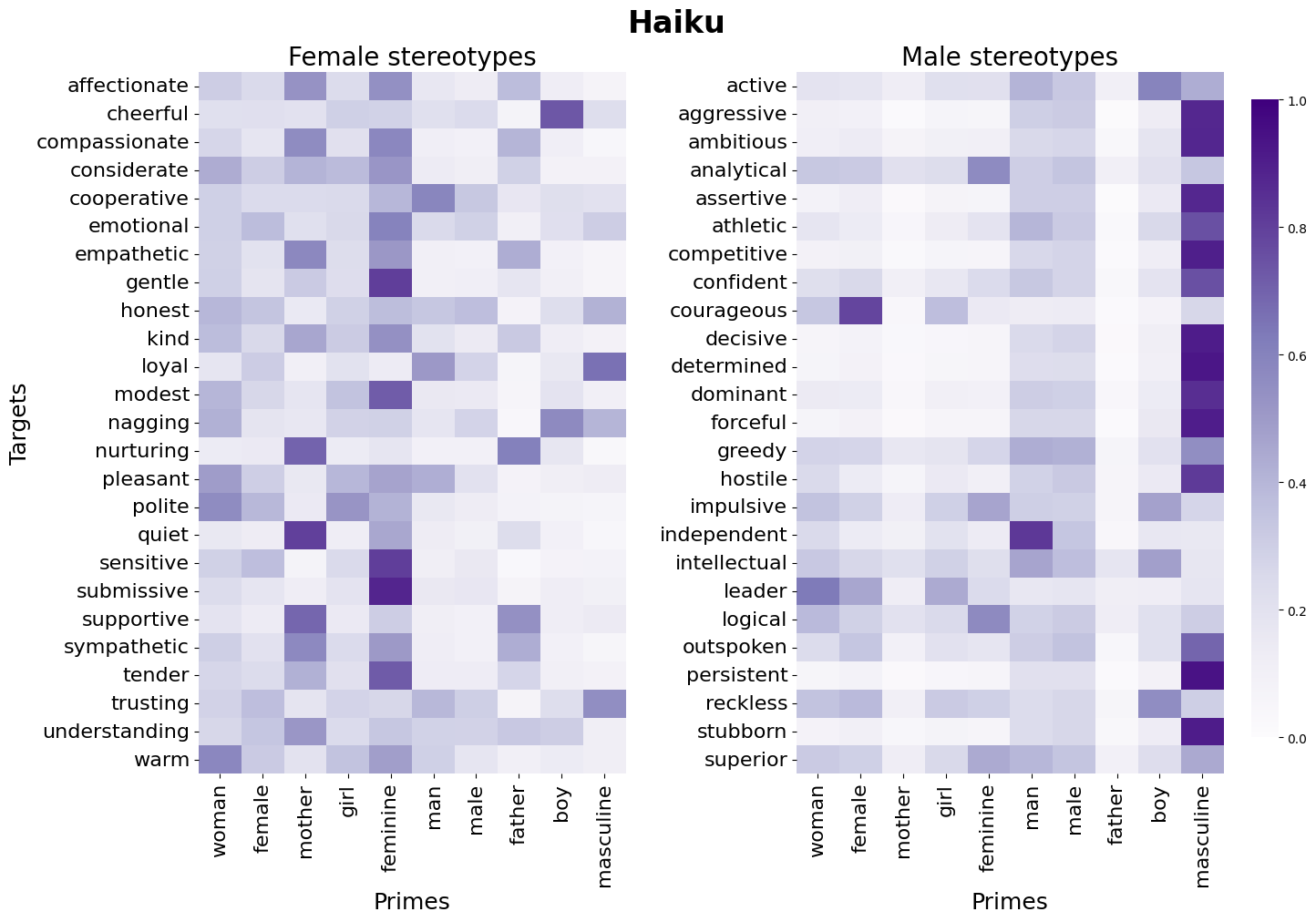}
\caption{\textbf{Gender biases in Haiku.} The heatmaps show the normalized final activation levels of 25 female-related targets (left) and 25 male-related targets (right) after activating 10 gender-related prime nodes in the Haiku network.}
\label{fig:heat_haiku}
\end{figure}

A fine-grained view of the results are shown in the heatmaps in Figures \ref{fig:heat_humans}, \ref{fig:heat_mistral}, \ref{fig:heat_llama3}, and \ref{fig:heat_haiku}, which show normalized final ALs of all prime-target pairs for each of the four networks, with female-related targets on the left and male-related targets on the right. The heatmaps make it easy to visually identify the presence of stereotype biases by observing a vertical divide in the lightness/darkness between the left and right sides of the heatmaps, corresponding to the female-related primes and the male-related primes. This is especially noticeable in the Llama3 heatmaps in Figure \ref{fig:heat_llama3}, indicating that male stereotypes (right) are stronger than female stereotypes (left), which is confirmed by the effect sizes. The heatmaps also make it possible to observe the strength of association between specific pairs of concepts, enabling a deeper understanding of what drives the stereotypes that are observed. For example, in all networks, the concept \textit{masculine} is strongly associated with many of the male-related targets, suggesting that the use of this term may have a stronger stereotypical connotation with respect to other male-related primes, such as \textit{father}, which appears to be weakly associated with many of the male-related targets. Also, some specific prime-target associations are strong across all networks, like \textit{mother -- nurturing}, but other associations differ greatly across the networks. Haiku, for instance, associates \textit{courageous} much more with \textit{female} than with \textit{male}, a stereotype-inconsistent pattern that stands out compared to the other networks.

\begin{figure}[H]
\centering
\includegraphics[width=.85\linewidth]{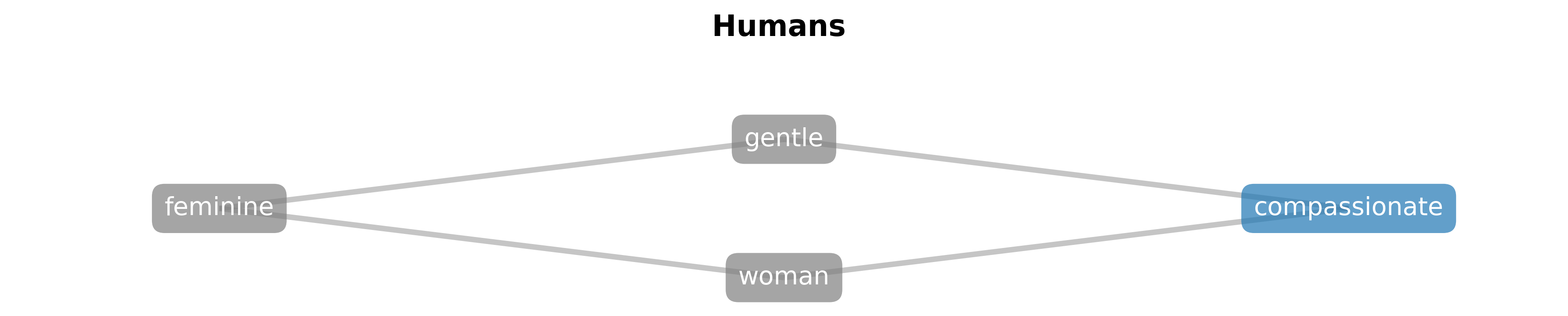}
\smallskip
\includegraphics[width=1\linewidth]{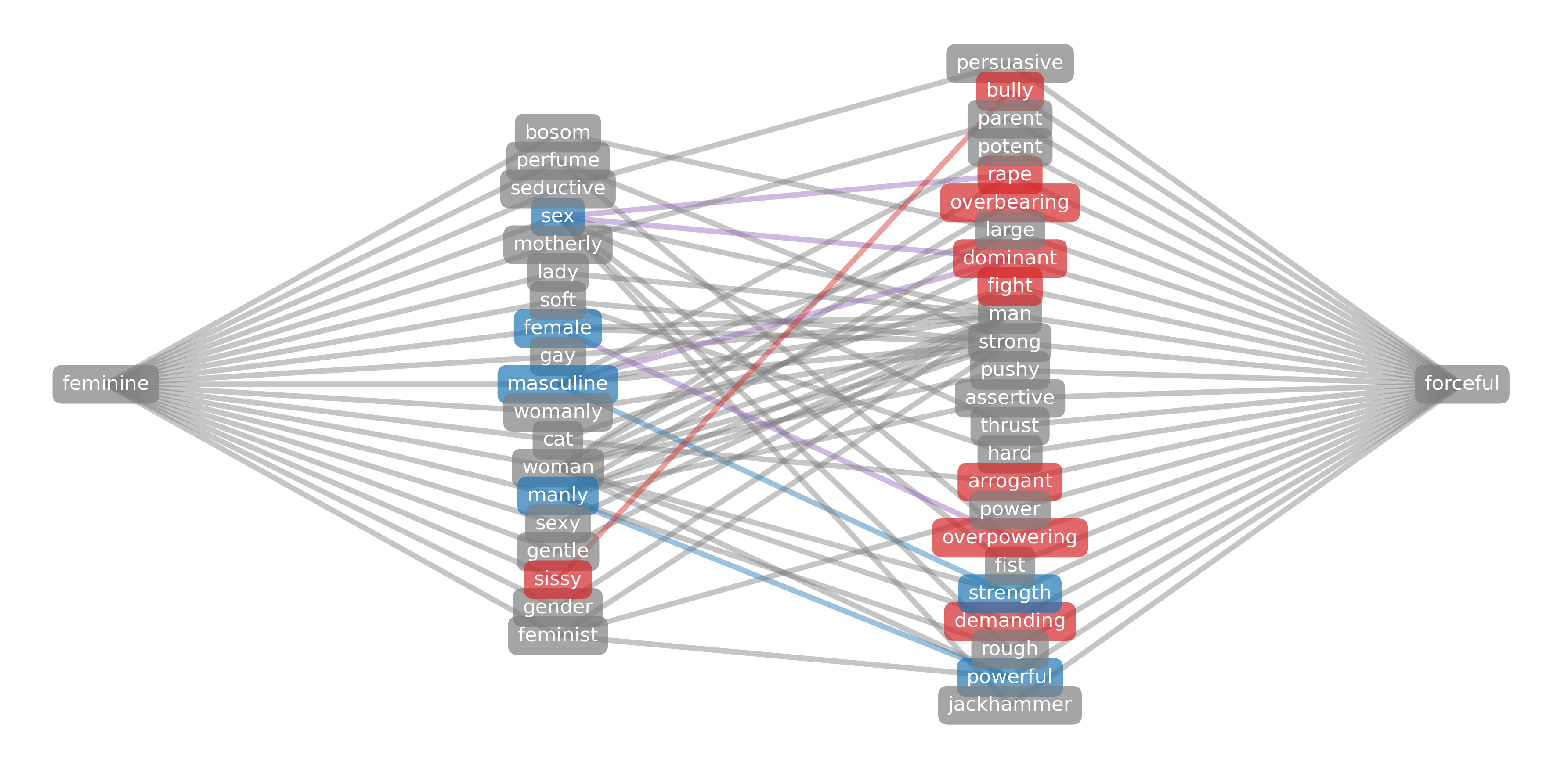}
\includegraphics[width=.85\linewidth]{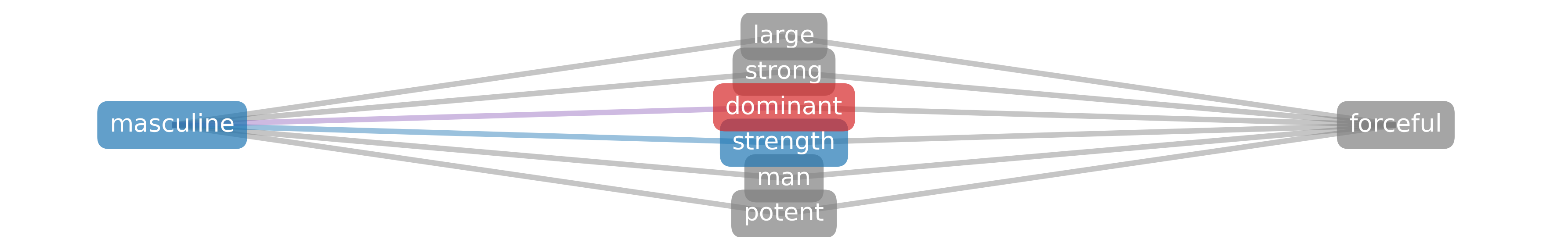}
\includegraphics[width=1\linewidth]{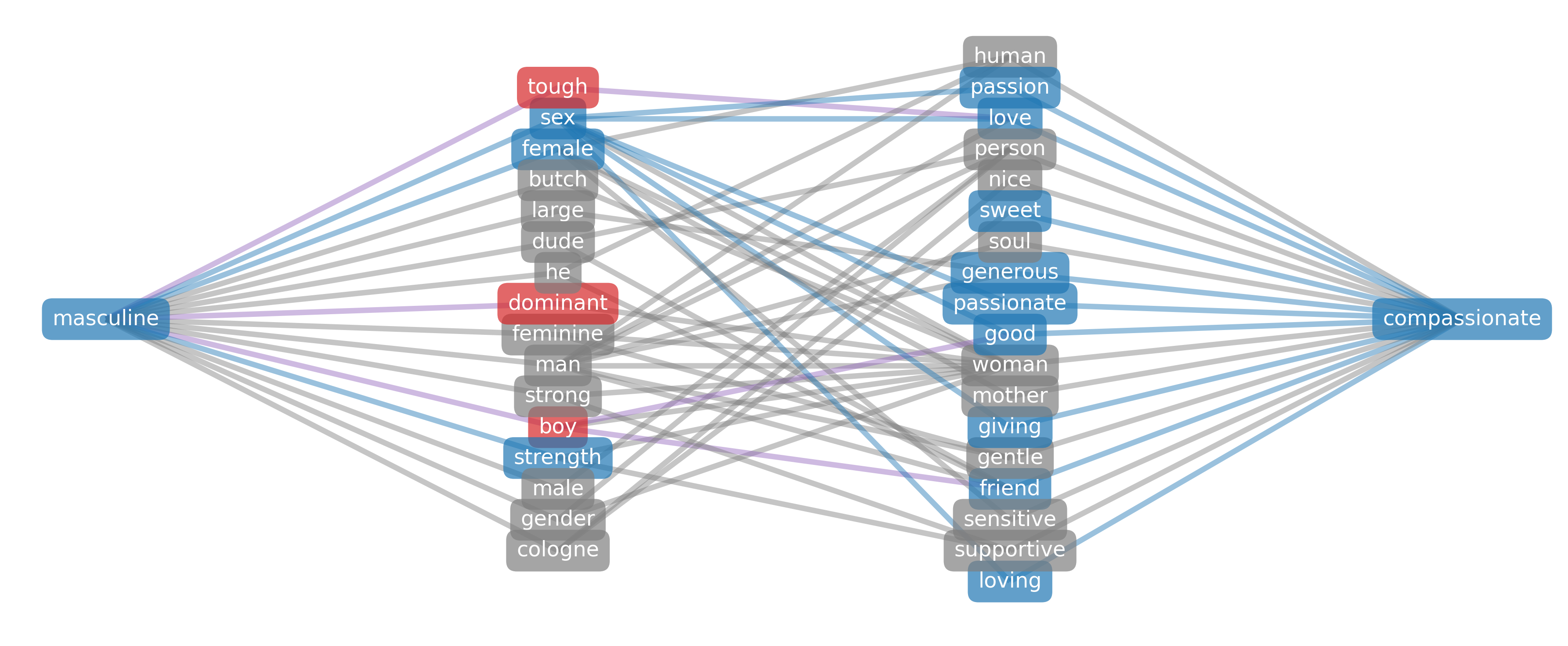}
\caption{\textbf{Human mindset streams.} The mindset streams show the paths from primes to targets in the human network for four different prime-target pairs. The colors of the words reflect human valence ratings of the words. Paths are typically shorter for stereotype-consistent pairs (\textit{feminine -- compassionate}, \textit{masculine -- forceful}) compared to stereotype-inconsistent pairs (\textit{feminine -- forceful}, \textit{masculine -- compassionate}).}
\label{fig:humans_mindset}
\end{figure}

\begin{figure}[H]
\centering
\includegraphics[width=.85\linewidth]{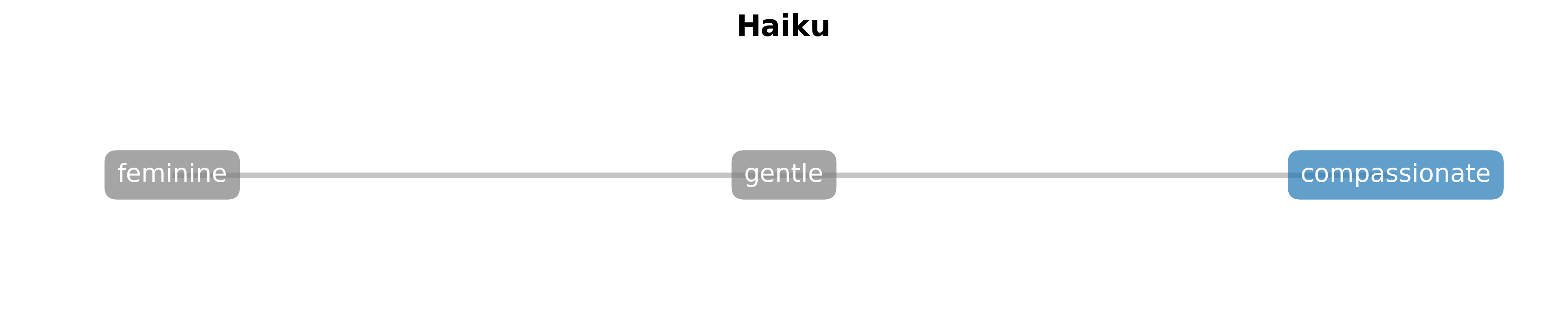}
\smallskip
\includegraphics[width=.9\linewidth]{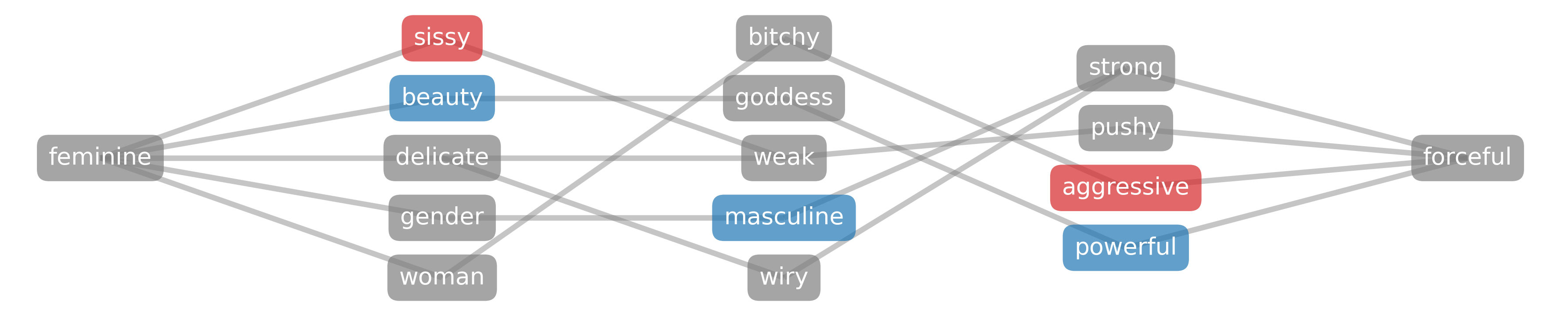}
\includegraphics[width=.85\linewidth]{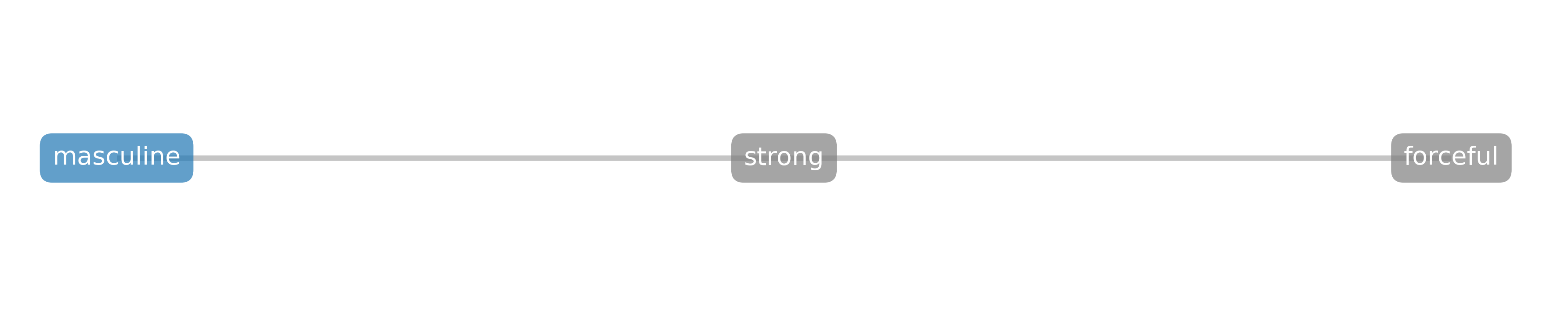}
\includegraphics[width=.9\linewidth]{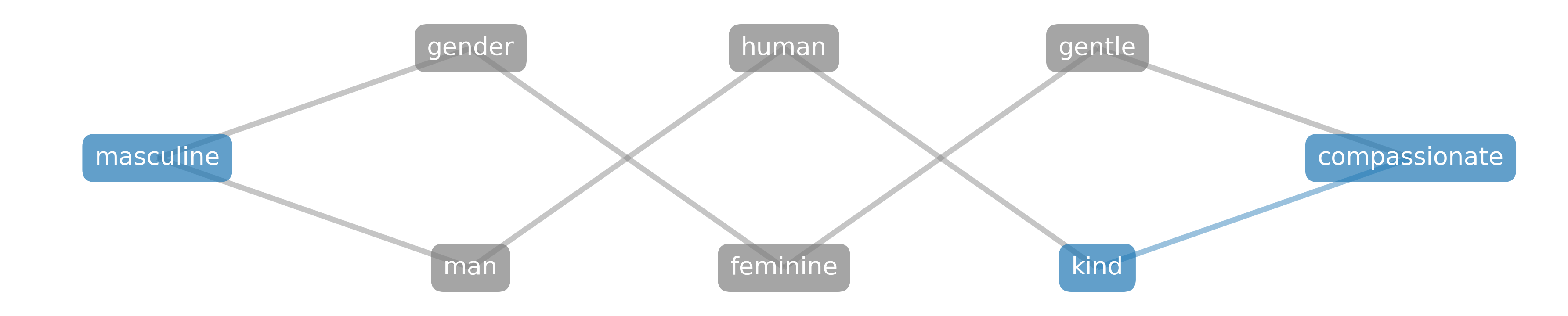}
\caption{\textbf{Haiku mindset streams.} The mindset streams show the paths from primes to targets in the Haiku network for four different prime-target pairs. The colors of the words reflect human valence ratings of the words. Paths are typically shorter for stereotype-consistent pairs (\textit{feminine -- compassionate}, \textit{masculine -- forceful}) compared to stereotype-inconsistent pairs (\textit{feminine -- forceful}, \textit{masculine -- compassionate}).}
\label{fig:haiku_mindset}
\end{figure}

A deeper qualitative understanding of stereotype biases can be gained by the observing paths between specific prime-targets pairs, as shown in Figures \ref{fig:humans_mindset} and \ref{fig:haiku_mindset}. We call these paths "mindset streams" because they show the stream of thought that is traveled in implicit semantic memory to get from one concept to another. Figures \ref{fig:humans_mindset} and \ref{fig:haiku_mindset} show mindset streams for humans and Haiku respectively, for four prime-target pairs: \textit{feminine -- compassionate},  \textit{feminine -- forceful}, \textit{masculine -- forceful}, and \textit{masculine -- compassionate}. One of the first things we note is that paths between stereotype-consistent pairs are typically shorter than those between stereotype-inconsistent pairs. We observe this in both humans and Haiku. The paths from \textit{feminine} to \textit{forceful} and from \textit{masculine} to \textit{compassionate} (stereotype-inconsistent) are longer and contain more nodes along the way. The nodes along the way provide important insight about the contexts through which the primes and targets are associated. The mindset streams are enriched with valence data from human ratings, so positive words are colored blue while negative words are colored red. These colors make it easy to identify certain patterns of positive or negative attitudes within the stereotypes (not to be confused with the results from the valence approach that we discuss later). For example, in the human mindset streams from \textit{feminine} to \textit{forceful} and from \textit{masculine} to \textit{compassionate}, which are stereotype-inconsistent, we observe many nodes along the paths that are related to the concept \textit{sex}, while such nodes are completely absent in the stereotype-consistent paths since the associations are not mediated by the context of sex. There are also more negative concepts than positive concepts along the path from \textit{feminine} to \textit{forceful} while there are more positive concepts than negative concepts along the path from \textit{masculine} to \textit{compassionate}. This suggests that going against the female stereotype may be viewed more negatively than going against the male stereotype, which has important implications for sexism and the struggle for gender equality. The mindset streams also highlight important differences between humans and Haiku. We observe that the paths are much richer in the human network than in the Haiku network, which we would expect given the network statistics (many more nodes and edges in the human network). Thus, Haiku displays a much more simplistic and limited conceptual knowledge structure compared to humans. In fact, the stereotype-inconsistent paths in the Haiku network are even longer than those in the human network. Similar to humans, we observe more negative concepts along the paths from \textit{feminine} to \textit{forceful} than those from \textit{masculine} to \textit{compassionate}. However, unlike in humans, the context of sex does not mediate the association between stereotype-inconsistent pairs in Haiku. The effect sizes, together with the heatmaps and the mindset streams provide a comprehensive and varied understanding of stereotype biases in humans and LLMs, allowing for many different types of comparisons and insights.

\subsubsection*{Valence related to religion, ethnicity, and sexual orientation
}
\begin{figure}[H]
\centering
\includegraphics[width=1\linewidth]{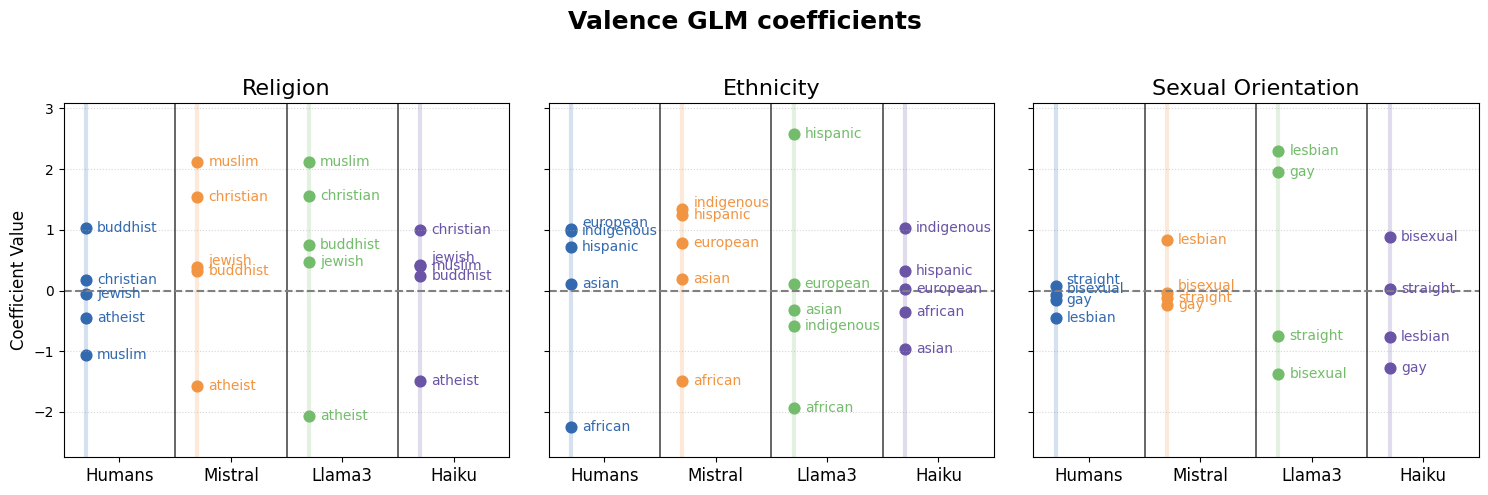}

\caption{\textbf{Coefficients reflecting positive and negative perceptions.} The plots show coefficient values of the separate univariate GLMs fit for each category of religion (left), ethnicity (center), and sexual orientation (right) for each of the four networks. Positive coefficients indicate that the category is perceived more positively, while  negative coefficients indicate that the category is perceived more negatively, with respect to the other categories.}
\label{fig:valence_coeff}
\end{figure}

\begin{figure}[H]
\centering
\includegraphics[width=1\linewidth]{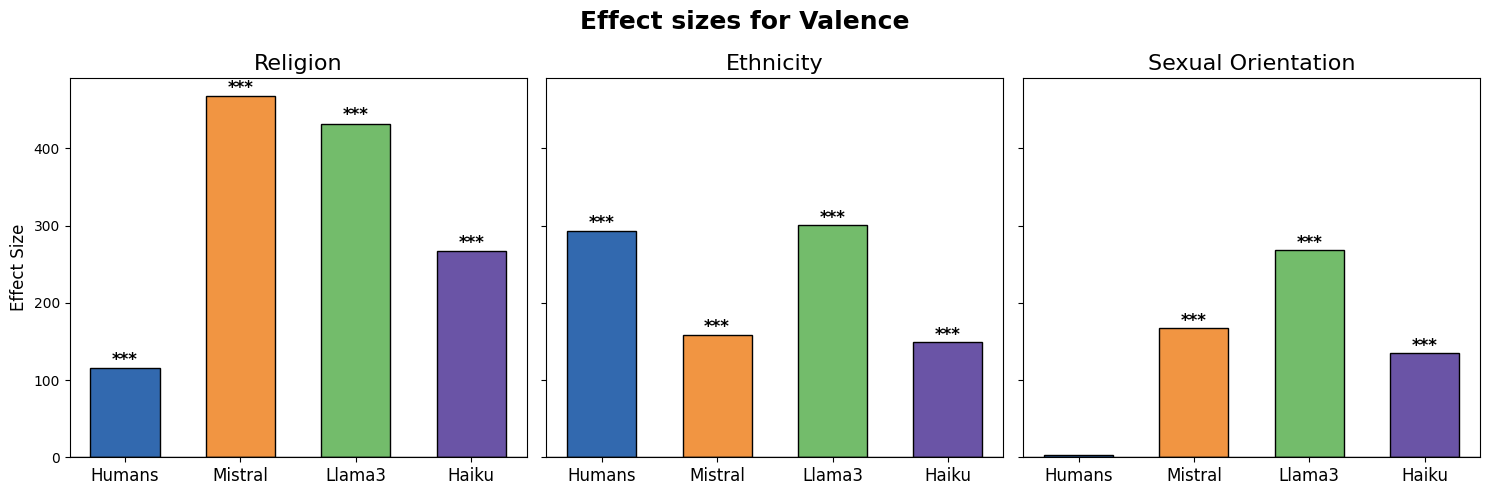}
\caption{\textbf{Effect sizes for valence GLMs.} The barplots show the effect sizes for each network for each religion (left), ethnicity (center), and sexual orientation (right), computed using the Wald test for a single multivariate GLM. Statistical significance is indicated with stars. A large effect size indicates that the coefficients of the GLM differ significantly from each other, indicating that there is a significant difference in how the different categories are perceived in terms of valence.}
\label{fig:valence_effects}
\end{figure}

We applied the valence approach for evaluating bias related to religion, ethnicity, and sexual orientation in humans, Mistral, Llama3, and Haiku. We implemented spreading activation processes in word association networks generated by humans and the three LLMs by activating prime nodes related to the three social identities in each of the four networks. Then, for each of the three social identities, we fitted separate univariate GLMs for each category (prime node) on the vectors of final ALs of all nodes such that the final ALs were the predictors and the valence scores of all the nodes were the responses. The coefficient values of each GLM indicate the extent to which the category (prime node) of the social identity is viewed positively or negatively with respect to the other categories. Effect sizes were computed by fitting a single multivariate GLM including the final AL vectors of all the categories as predictors and the valence scores as the response. Significant effect sizes indicate that there is large variation in the perception of the different categories (i.e. the coefficients differ significantly from each other). The coefficients of the separate univariate GLMs are shown for religion, ethnicity, and sexual orientation in Figure \ref{fig:valence_coeff} while the effect sizes computed from the single multivariate GLM are shown in Figure \ref{fig:valence_effects}. For religion, and ethnicity, there is significant valence bias in humans and all three LLMs. Instead, for sexual orientation, humans demonstrate almost no valence bias but all three LLMs demonstrate bias. Contrary to the gender stereotype biases, humans are less biased than the LLMs regarding religion, but regarding ethnicity, humans are just as biased if not more biased than the LLMs. 
For religion, \textit{buddhist} and \textit{christian} are seen more positively by humans while \textit{athiest} and \textit{muslim} are seen more negatively, and \textit{jewish} is seen as neutral. Interestingly, in all three LLMs, \textit{muslim} is seen more positively, and \textit{buddhist} is seen less positively among LLMs, contrary to humans. Regarding ethnicity, humans view \textit{african} much more negatively than the other ethnicities, and they view \textit{european} more positively, reflecting the bias of white supremacy. In LLMs, \textit{european} is not the most positively viewed ethnicity, but \textit{african} is more negatively viewed in all LLMs. Regarding sexual orientation, while humans demonstrate little to no bias, Llama3 views lesbians and gays more positively but bisexuals more negatively. The valence approach is mainly useful for observing how positive and negative perceptions differ across certain categories of a certain social identity, and it allows for comparisons between humans and LLMs. These comparisons are insightful, especially when the LLMs demonstrate patterns of perception that are contradictory to those of humans, because it suggests that maybe the LLMs were fine-tuned to correct certain biases, making them biased in the opposite direction.

\subsubsection*{Emotions related to political party
}
\begin{figure}[H]
\centering
\includegraphics[width=.8\linewidth]{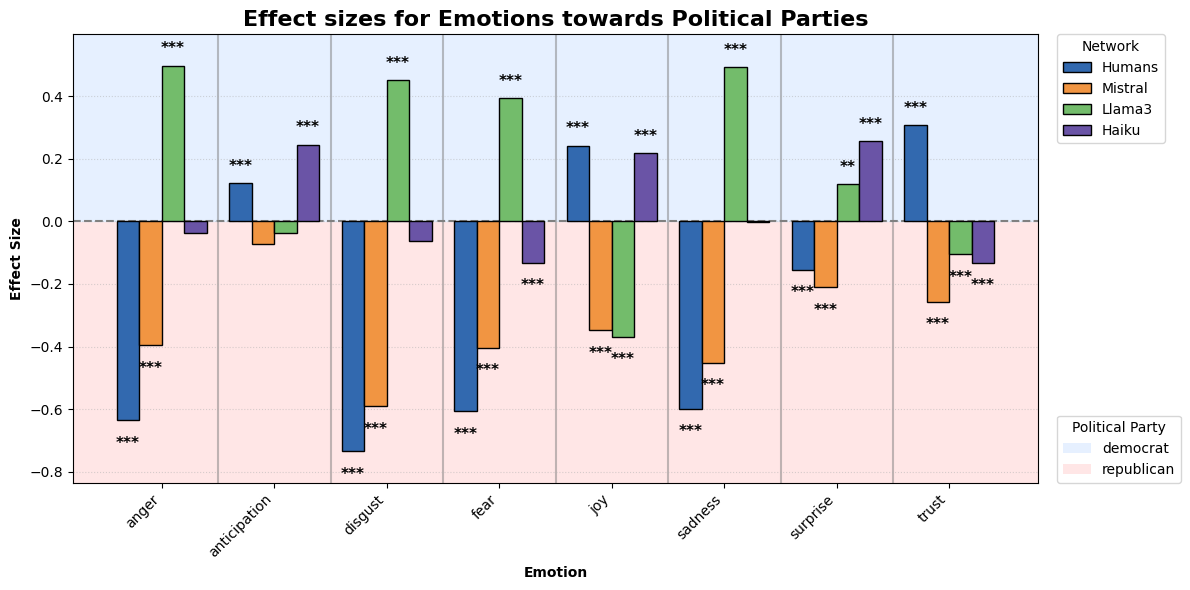}
\caption{\textbf{Effect sizes for emotions towards political parties.} The barplots show the effect sizes for each network for all eight core emotions. Statistical significance is indicated with stars. For each emotion, positive effects indicate stronger feelings of that emotion towards Democrats, while negative effects indicate stronger feelings towards Republicans. Note, the final activation level matrices were normalized using the L1 norm (see Methods).}
\label{fig:emotions}
\end{figure}

We applied the emotions approach for evaluating bias related to political party in humans, Mistral, Llama3, and Haiku. We implemented spreading activation processes in word association networks generated by humans and the three LLMs by activating the prime nodes \textit{democrat} and \textit{republican} in each of the four networks, and then we observed the final ALs of words that elicit certain emotions. We quantified relative differences in emotions felt towards Democrats and Republicans by computing differences in the final ALs of emotion-related target nodes for eight different emotions. The Wilcoxon test for paired samples was applied to the differences and effect sizes were computed for each emotion. The effect sizes are shown in the barplots in Figure \ref{fig:emotions}. Positive effect sizes indicate that a given emotion is felt more strongly towards Democrats, while negative effect sizes indicate that a given emotion is felt more strongly towards Republicans. Perhaps the first thing we notice is that the negative emotions of anger, disgust, fear, and sadness are felt more for Democrats in Llama3 but more for Republicans in humans and Mistral. Haiku demonstrates a more neutral stance for these four negative emotions, with only significant negative effects for fear (more strongly felt towards Republicans). Instead, for the positive emotions of anticipation, joy, surprise, and trust, we observe varying patterns. Notably, effect sizes for humans and LLama3 are of opposite polarity for all emotions. It is also interesting to note that only humans feel more trust towards Democrats, with all three LLMs having more trust in Republicans. It is also interesting to note that Mistral demonstrates stronger emotions for Republicans across all eight emotions, though not all of these effects are significant. Of all the networks, Haiku appears to be the most politically neutral in terms of emotions. On the contrary, humans appear to be the most radical, demonstrating especially strong emotional differences for negative emotions towards Republicans. The implications of these results are further elaborated in the Discussion.

\section{Discussion}
\label{sec:discussion}

In this work, we introduce a novel word association network methodology for evaluating implicit biases in LLMs compared to humans. We demonstrate the utility of our methodology by applying it to studying social biases in humans and three LLMs. We apply three different versions of our methodology to investigate different types of biases: a stereotypes approach to investigate stereotypical gender biases, a valence approach to investigate positive and negative perceptions of religion, ethnicity, and sexual orientation, and finally, an emotions approach to investigate the relative feelings towards Democrats and Republicans for eight core emotions. In what follows we discuss the results of our analyses, which indicate both convergence and divergence in biases between humans and LLMs, and we also discuss the advantages and societal implications of our methodology, as well as limitations and future work.

\subsection*{Evaluation of biases: stereotypes, valence, and emotions}
Convergence was most evident regarding gender stereotypes, in which stronger associations were observed between stereotypical prime-target pairs like \textit{feminine -- compassionate} and \textit{masculine -- forceful} for both humans and LLMs, consistent with widely documented linguistic and cultural stereotypes. Humans displayed the strongest overall gender bias, with equally strong female-related and male-related stereotypes, while the LLMs varied in both direction and intensity. Specifically, Mistral and Haiku showed stronger female-related stereotypes, whereas Llama3 exhibited the opposite pattern, demonstrating stronger male-related stereotypes. These results support previous work showing that LLMs can inherit human-like stereotypes from training data \cite{stella2023using,bai2025explicitly,caliskan2017semantics}, but also suggest that different models may internalize and express these associations differently depending on their training and fine-tuning procedures. The qualitative mindset streams further reveal that stereotype-consistent associations are not only stronger but also cognitively more accessible, as evidenced by shorter paths between primes and targets. This observation resonates with cognitive psychology findings that implicit associations are often more automatic and contextually reinforced \cite{greenwald1998measuring, wilson2000model}. The human mindset streams were also enriched with more nuanced associative contexts and valence variations, consistent with richer conceptual networks and more complex socially embedded knowledge structures. In contrast, Haiku’s pathways reflect a more simplistic conceptual architecture, suggesting a shallower representation of the complex, culturally mediated nuances that characterize gendered cognition in humans. Notably, the finding that stereotype-inconsistent associations (e.g. \textit{feminine -- forceful}) are mediated by more negative intermediate nodes in humans but not in Haiku suggests a crucial distinction: human biases often entail evaluative and affective components, whereas LLM biases are largely structural and probabilistic in nature.

The valence analysis further showed systematic differences in how social categories of varying religion, ethnicity, or sexual orientation are perceived. The finding that \textit{christian} elicited more positive associations than \textit{muslim} is consistent with IAT research showing pro-Christian implicit preferences in Western samples and broader implicit hierarchies across religions \cite{heiphetz2013patterns}. Interestingly, all three LLMs also perceived \textit{christian} positively, but Mistral and Llama3 perceived \textit{muslim} even more positively than \textit{christian}, in contrast to humans. Alignment between LLMs and humans may suggest that LLMs not only replicate but may amplify valence asymmetries in ways that risk reinforcing social inequalities, as shown in previous work on math anxiety in LLMs \cite{abramski2023cognitive}. However, misalignment between LLMs and humans may reflect efforts to reduce or counterbalance biases in their fine-tuning data \cite{shen2023large}. Ethnicity-related biases were also consistent with prior sociolinguistic findings -- humans demonstrated clear patterns of ingroup favoritism, with \textit{european} perceived most positively and \textit{african} most negatively, reflecting structural inequalities in Western corpora and cognition \cite{acerbi2023large}. The LLMs mirrored this pattern to varying degrees, with \textit{african} consistently associated with lower valence, suggesting that despite extensive filtering and alignment, models continue to internalize biases present in large-scale English-language training data \cite{bender2021dangers, sorokovikova2025surface}. Results regarding sexual orientation were particularly revealing -- humans showed negligible bias, while all three LLMs exhibited significant differences in perception, often favoring gay and lesbian identities. These findings may reflect a growing over-correction effect in alignment pipelines, where biases are mitigated in one direction but inadvertently amplified in another \cite{sun2025large}. Collectively, these results highlight that while LLMs often reproduce human-like biases, they can also develop “counter-biases” that diverge systematically from human cognitive tendencies. Moreover, evidence that LLMs reproduce culturally prevalent content biases suggests that unequal valence may reflect corpus-level regularities rather than purely model-internal artifacts \cite{acerbi2023large}. Future work should test whether targeted corpus curation or debiasing alters these valence profiles and whether network-level edits can attenuate amplification effects without degrading linguistic competence.

Regarding emotions in the political domain, significant divergences between humans and LLMs emerged. Humans exhibited strong emotional polarization, with negative emotions such as anger, disgust, fear, and sadness directed primarily toward Republicans, and positive emotions such as trust and joy toward Democrats. These patterns align with evidence of affective polarization in U.S. populations \cite{stanley2011implicit, sun2025large}. In contrast, the LLMs demonstrated less emotional differentiation overall, with Mistral and Llama3 showing mild polarity and Haiku appearing largely neutral. Interestingly, all three models expressed more trust toward Republicans, opposite to the human pattern. These discrepancies highlight that while structural features of bias resemble human cognition, the content of associations may diverge in politically sensitive domains. Such divergences may reflect differences in cultural representation within training corpora \cite{stanley2011implicit,stella2023using,acerbi2023large} or limitations in how LLMs express emotional content \cite{sun2025large}. Recent audits report systematic political leanings in several LLMs, which could partially explain model–human mismatches in affective attributions \cite{motoki2024morehuman,rozado2024political}. Moreover, political preferences in LLMs have been shown to shift with prompts, personas, and scale, underscoring the sensitivity of affective associations to elicitation conditions \cite{bang2024measuring,motoki2024morehuman}. Independent evidence also indicates that current systems exhibit uneven emotional intelligence and gaps in emotion reasoning, consistent with the affective divergences observed here \cite{abramski2023cognitive,sun2025large}. Together, these findings support the need for interpretable, network-based diagnostics that can localize which subgraphs or pathways (e.g. mindset streams \cite{brian2023introducing}) encode partisan-affective couplings rather than relying solely on aggregate bias scores.

\subsection*{Methodological advantages and societal implications}
A central methodological innovation of the present work is the use of word association networks \cite{abramski2025llm} to directly compare implicit biases in humans and LLMs. Unlike model-level embedding-based metrics, which often operate as opaque vector-space measures \cite{caliskan2017semantics}, our network-based methodology does not require access to internal representations, rather, it offers interpretable structures in which nodes represent concepts and edges capture associative strength \cite{semeraro2025emoatlas,siew2019cognitive,brian2023introducing}. Moreover, unlike most output-level metrics that use prompts to evaluate bias in specific contexts, our methodology captures implicit relational knowledge rather than surface-level outputs. One of the main advantages of our methodology is that it is grounded in cognitive psychology and network science. Semantic networks of memory recall patterns have been widely applied to model conceptual memory \cite{de2019small}, spreading activation \cite{citraro2025spreadpy}, and semantic priming \cite{siew2019spreadr} in humans. By operationalizing implicit bias as asymmetries in activation flow across these networks, our approach mirrors established theories of semantic priming \cite{collins1975spreading,siew2019spreadr} while remaining transparent and human-readable. The networks constructed in this study capture large-scale associative structures. For example, the human network consists of 24,308 nodes and 317,344 edges, while the Llama3 network is even larger with 38,987 nodes and 546,866 edges, reflecting differences in representational richness across agents. This transparency allows biases to be traced through activation pathways, (i.e. mindset streams) \cite{brian2023introducing}, such as the observation that stereotype-consistent links (e.g. \textit{masculine -- forceful}) are mediated by shorter semantic paths than inconsistent ones. 
The interpretability afforded by network-based approaches therefore provides a clear advantage, as it enables both quantitative evaluation and qualitative inspection of how associations emerge and propagate, supporting more nuanced accountability in bias evaluation \cite{abramski2023cognitive}.

To summarize, our word association network methodology offers three key contributions. First, it provides a flexible, scalable framework for implicit bias evaluation that avoids reliance on internal model representations, bridging the gap between model-level (intrinsic) and output-level (extrinsic) evaluations. Second, our approach is based on the IAT for assessing implicit biases in humans, and it is grounded in theories from cognitive science. And third, it enables direct quantitative and qualitative comparisons between humans and LLMs, a feature often missing from prior work in which claims about "human-like" LLM biases remain vague and ungrounded. This last point addresses the problematic practice of comparing LLM metrics to human metrics that are not obtained in the same way, such as comparing embedding-based results to IAT results, which can lead to unreliable comparisons \cite{bai2025explicitly}. To the best of our knowledge, our methodology is the first of its kind that offers both quantitative and qualitative assessments of implicit bias enabling direct comparisons between humans and LLMs.

The societal implications of adopting a word association network approach to LLM bias evaluation are many. Because word association networks provide interpretable pathways of associations \cite{semeraro2025emoatlas}, they allow stakeholders—including policymakers, educators, and developers—to identify not only the presence of bias but also the mechanisms through which it propagates \cite{abramski2023cognitive}. This level of transparency is crucial in high-stakes domains such as healthcare, hiring, and criminal justice, where biased outputs from LLMs can perpetuate discrimination \cite{motoki2024morehuman} or reinforce social stereotypes \cite{caliskan2017semantics,sun2025large}. By linking model associations to human cognitive structures, this method enables a clearer assessment of whether LLMs amplify existing human biases or introduce qualitatively different patterns, which has direct implications for trust and accountability in AI systems. Furthermore, the network framework supports cross-cultural and longitudinal monitoring of bias, making it possible to track how implicit associations evolve as models are updated or deployed in new contexts \cite{abramski2024llm,abramski2025llm}. In doing so, it provides a scalable tool for guiding socially responsible AI development, fostering greater alignment between technological innovation and ethical standards in society.

\subsection*{Limitations and future work}

Despite the many advantages of our word association network methodology, it also has several limitations that should be taken into careful consideration and that could be further addressed in future work. One important limitation has to do with the normalization of the final activation level matrices, which is necessary in order to account for differences in the centrality of nodes within the word association networks. This normalization requires choosing the appropriate norm, but the choice can potentially influence the results produced by the methodology. In this work, we focused on two highly common norming choices -- the L1 norm and the L2 norm. For the stereotypes approach and the valence approach, the L1 norm and the L2 norm produced results that converged towards the same patterns, however, for the emotions approach, we found that the selection of the L1 norm (see results in Figure \ref{fig:emotions}) led to different results compared to the L2 norm (see results in Figure \ref{fig:emotions2} in the Appendix). This variability in results suggests that the methodology entails a level of complexity that requires an accurate selection of the norm by the researcher. In the case of the analysis of feelings towards Democrats and Republicans, selecting the L2 norm resulted in effect sizes of the same polarity across all emotions for a given network, for example, humans displaying stronger emotions towards Republicans for all eight emotions, as shown in Figure \ref{fig:emotions2}. Such a result is odd because it means that contradictory emotions, for example, joy and sadness, are simultaneously felt for the same political party. Instead, according to Plutchik's theory of emotions \cite{plutchik1980general}, we would expect contradictory emotions to be felt more strongly towards opposite political parties, for example, joy towards Democrats and sadness towards Republicans, which is what we observe in humans when we use the L1 norm, as shown in Figure \ref{fig:emotions}. Thus, we selected the L1 norm because the results it produced align more closely with the pattern we would expect to observe according to Plutchik's theory. We argue that this case study using the emotions approach is interesting as it underlines the need for the researcher to perform appropriate parameter selection based on relevant theoretical knowledge. A key limitation of our approach is that, depending on the location of words within the networks, the L1 and L2 norms might produce different results. Thus, the norm should be selected by the researcher upon careful consideration and interpretation of the results.

Our methodology features additional limitations. First, the word association networks do not have multiple nodes for words with multiple senses, and so words like \textit{black} and \textit{white}, which could potentially be chosen as prime nodes for investigating racial biases, can also be interpreted as colors. Therefore, results may be influenced by confounding when words with multiple meanings are included as prime nodes or target nodes. Second, the reliance on free-association norms constrains the ecological validity of the findings. While word association networks provide a valuable proxy for implicit cognitive structures \cite{de2019small,siew2019cognitive,citraro2025spreadpy}, it remains uncertain to what extent they predict real-world behaviors in domains such as healthcare, recruitment, or criminal justice, where implicit bias has direct social consequences \cite{zhao2025explicit,perugini2005predictive}. However, it should be noted that this limitation is not unique to our methodology. In fact, the extent to which human IAT results can be used to predict discriminatory behavior in applied settings remains a topic of debate as well \cite{schutzle2025does}. Implicit bias evaluation provides insight into latent representational structures that influence thinking and reasoning, but decision-making in  applied settings involves contextual and normative considerations that cannot easily be captured by intrinsic measures. Third, the SWOW dataset of free association norms \cite{de2019small} used to build the human word association network includes mostly participants whose native language is English (i.e. the US and the UK) and so the implicit knowledge that is modeled by the human network reflects a Western-centric worldview. As previous work has demonstrated \cite{xiang2025comparing}, cultural and linguistic contexts strongly shape the structure of cognitive memory networks, meaning that implicit associations in multilingual or non-Western settings may diverge significantly from those represented here. Future work should aim to use multilingual or cross-cultural human-generated data to provide a more globally representative human benchmark, though such datasets are limited. Fourth, in this study, only three LLMs (Mistral, Llama3, and Haiku) were evaluated. Although these LLMs differ in scale and architecture, this small sample limits generalization to the broader ecosystem of models, including those with distinct training pipelines (e.g. models suitable for cognitive science experiments \cite{binz2025foundation}). Finally, both human and machine biases are dynamic rather than static \cite{guo2024bias}. Human associative norms evolve over time, and LLMs are continuously updated with new training data and alignment strategies, which can shift their internal representational structures as models evolve. The importance of this evolutionary component in the context of LLMs is highlighted by Abramski et al. \cite{abramski2023cognitive}, who found that negative perceptions of STEM subjects were progressively reduced in newer versions of GPT models. Longitudinal approaches and cross-cultural replications \cite{acerbi2023large} will therefore be essential for determining the robustness of the present findings. Future research should broaden the scope by applying this methodology to larger and more diverse sets of models and cultural contexts, while triangulating free-association data with behavioral and decision-level measures of bias.

\subsection*{Conclusions}
In this work, we introduce a novel methodology for evaluating and interpreting implicit biases in LLMs with respect to humans based on word association networks. Our approach, which aligns with the implicit association test (IAT) for measuring biases in humans, offers a cognitively grounded, scalable, and transparent framework for assessing biases both quantitatively and qualitatively. We apply our methodology to investigate implicit social biases using three different approaches that assess various ways in which biases can be expressed: through stereotypes, valence, and emotions. Our results reveal that implicit biases in LLMs, while structurally similar to those of humans, diverge in content, polarity, and intensity across various social domains. Our findings underscore the need for systematic methodologies that can capture both parallels and mismatches between human and machine associative structures. Beyond bias evaluation, our approach contributes to the broader goal of guiding the socially responsible design and deployment of language technologies that have a positive impact on society.



\bibliography{main}



\section*{Author contributions statement}
K.A., G.R., and M.S. conceived the methodology. K.A. conducted the analyses and analyzed the results. K.A., G.R., and M.S. interpreted the results and reviewed the manuscript.

\section*{Acknowledgments}
This work is supported by: (i) SoBigData.it which receives funding from the European Union -- NextGenerationEU -- National Recovery and Resilience Plan (PNRR); (ii) the Future Artificial Intelligence Research (FAIR) Foundation, a non-profit foundation financed under the National Recovery and Resilience Plan (PNRR).

\section*{Competing interests} 
The authors declare no competing interests.

\section*{Appendix} 
\begin{figure}[H]
\centering
\includegraphics[width=.8\linewidth]{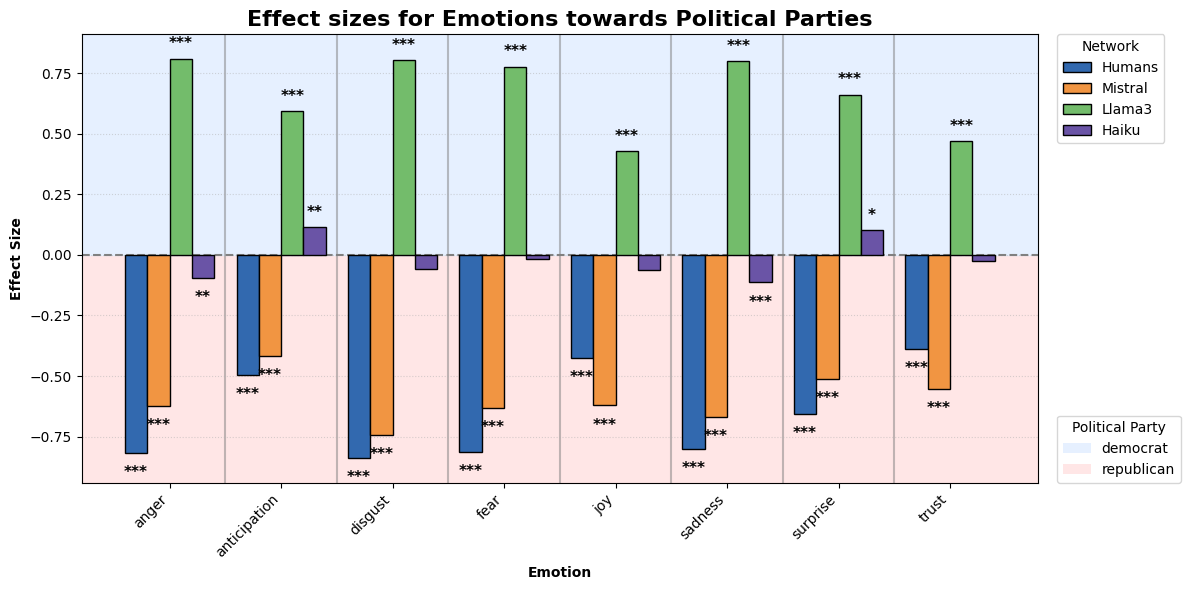}
\caption{\textbf{Effect sizes for the emotional analysis of political parties (L2 normalization).} The barplots show the effect sizes for each network for all eight core emotions. Statistical significance is indicated with stars. For each emotion, positive effects indicate stronger feelings of that emotion towards Democrats, while negative effects indicate stronger feelings towards Republicans. Note, the final activation level matrices were normalized using the L2 norm (see Methods).}
\label{fig:emotions2}
\end{figure}

\end{document}